\RequirePackage[loading]{tracefnt}
\documentclass[lettersize,journal]{IEEEtran}
\usepackage{times}
\usepackage{multicol}
\usepackage[bookmarks=true]{hyperref}
\usepackage{graphicx}
\usepackage{times}
\usepackage{graphicx}
\usepackage{booktabs}
\usepackage{amsmath}
\usepackage{biblatex}
\usepackage{subcaption}
\usepackage{mwe}
\usepackage{multirow}
\usepackage{wrapfig}
\usepackage{upgreek}
\usepackage{caption}
\usepackage{subcaption}
\usepackage{color}
\usepackage{amssymb}
\usepackage{comment}
\bibliography{references}

\graphicspath{ {./figures/} }

\begin{document}

\title{Design of an Adaptive Lightweight LiDAR to Decouple Robot-Camera Geometry} 

\author{Yuyang Chen$^{*1}$, Dingkang Wang$^{*2}$, Lenworth Thomas$^{3}$, Karthik Dantu$^{1}$, Sanjeev J. Koppal$^{2,4}$

\thanks{$^{1}$Yuyang Chen and Karthik Dantu are with the Department of Computer Science and Engineering, University at Buffalo, Buffalo, NY 14260, USA
        {\tt\small yuyangch@buffalo.edu, kdantu@buffalo.edu}}%
\thanks{$^{2, 3}$Dingkang Wang and Sanjeev Koppal are with the Department of Electrical and Computer Engineering; Lenworth Thomas is with the Department of Mechanical and Aerospace Engineering, University of Florida, Gainesville, FL 32603, USA
{\tt\small noplaxochia@ufl.edu, sjkoppal@ece.ufl.edu, lenworth.thomas@ufl.edu}}%

\thanks{$^{4}$
Sanjeev J. Koppal holds concurrent appointments as an Associate Professor of ECE at the University of Florida and as an Amazon Scholar at Amazon Robotics. This paper describes work performed at the University of Florida and is not associated with Amazon. }

\thanks{$^{*}$Equal contributions. }
}

\maketitle

\begin{abstract}
A fundamental challenge in robot perception is the coupling of the sensor pose and robot pose. This has led to research in active vision where robot pose is changed to reorient the sensor to areas of interest for perception. Further, egomotion such as jitter, and external effects such as wind and others affect perception requiring additional effort in software such as image stabilization. This effect is particularly pronounced in micro-air vehicles and micro-robots who typically are lighter and subject to larger jitter but do not have the computational capability to perform stabilization in real-time. We present a novel microelectromechanical (MEMS) mirror LiDAR system to change the field of view of the LiDAR independent of the robot motion. Our design has the potential for use on small, low-power systems where the expensive components of the LiDAR can be placed external to the small robot. We show the utility of our approach in simulation and on prototype hardware mounted on a UAV. We believe that this LiDAR and its compact movable scanning design provide mechanisms to decouple robot and sensor geometry allowing us to simplify robot perception. We also demonstrate examples of motion compensation using IMU and external odometry feedback in hardware. 

\end{abstract}

\IEEEpeerreviewmaketitle

\section{Introduction}

Modern autonomy is largely driven by vision and depth sensors for perception. Most such techniques make an implicit assumption that the relative pose of the sensor w.r.t. the robot is fixed and changes in sensor viewpoint require a change in the robot pose. This implies that fast-moving robots must deal with motion compensation (i.e. camera-robot \emph{stabilization}) and that robots need to reorient themselves to observe the relevant parts of the scene. Correspondingly, stabilization~\cite{neuhaus2018mc2slam,gojcic2019perfect,taketomi2017visual,phelps2019blind} and active vision~\cite{costante2016perception,bircher2016receding,zhang2018perception,sadat2014feature} are well-studied problems.

Let us consider the specific example of image stabilization. While successful, most such methods compensate through \emph{post-capture} processing of sensor data. \emph{We contend that this is simply not feasible for the next generation of fast miniature robots} such as robotic bees~\cite{robobees-sciam13}, crawling and walking robots \cite{hoffman2013design}, and other micro-air vehicles~\cite{mulgaonkar2015design}.  For example, flapping-wing robots such as the RoboBee exhibit a high frequency rocking motion (at about 120 Hz in one design) due to the piezo-electric actuation \cite{helbling2014pitch}. Environmental factors such as wind affects micro-robots to a greater extent than a larger robot. There might be aerodynamic instability due to ornithopter-based shock absorption \cite{xue2019computational}. The egomotion of small robots (and onboard sensors) is quite extreme making any sensing challenging. While there have been software methods to correct for such effects for cameras \cite{chen2009real} and LiDARs \cite{peng2007model}, this is often difficult to perform in real-time onboard due to the computational, energy  and latency constraints on the robot mentioned above. Without proper motion compensation for miniature devices, we will not be able to unlock the full potential of what is one of the ten grand challenges in robotics~\cite{yang2018grand}. 

\subsection{Key Idea: Compensation \emph{during} Imaging}

\begin{figure}
    \centering
    \includegraphics[width=0.45\textwidth]{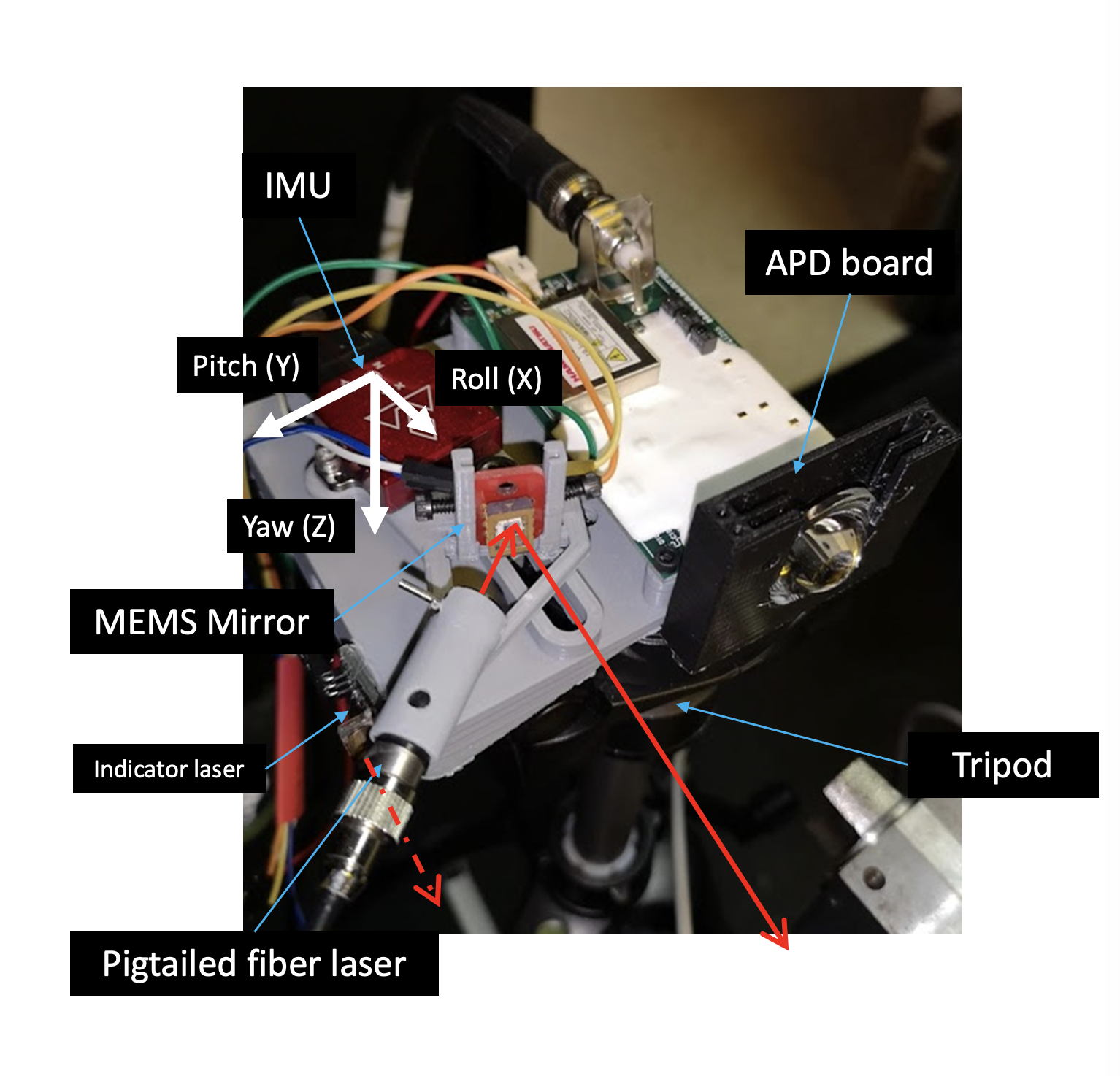}
    \includegraphics[width=0.2\textwidth]{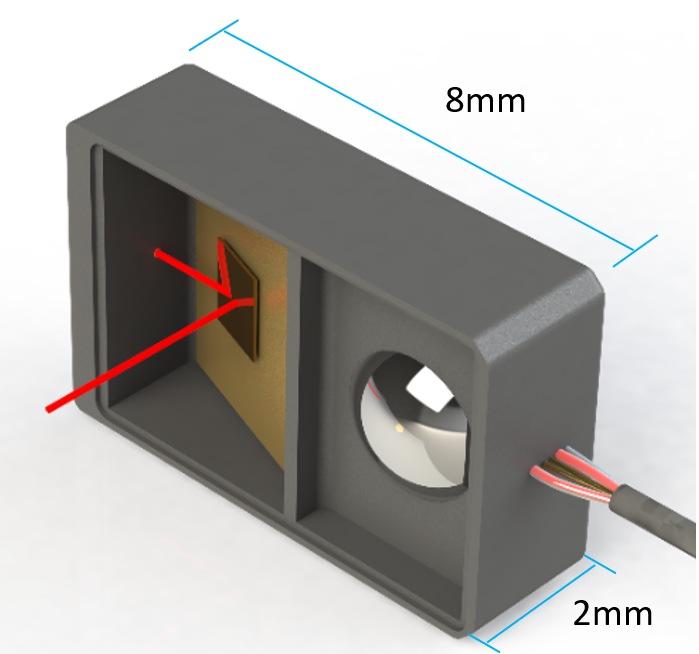}
    \includegraphics[width=0.2\textwidth]{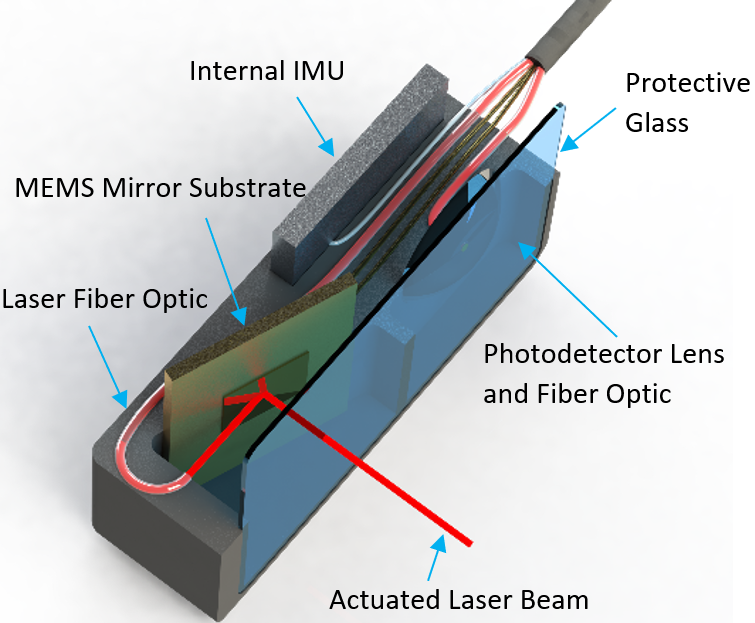}
    \caption{Our design is given above with the prototype motion-compensated LiDAR (up), and we also prepared a design for future work to integrate this onto smaller platforms.
    }
    \label{fig:futuristic}
    \vspace{-10pt}
\end{figure}
Our idea is for motion correction to happen in sensor hardware during imaging such that measurements are already compensated without requiring onboard computing. This paper shows the motion compensation advantage of decoupling robot-camera geometry, and providing the ability to control the camera properties independent of the robot pose could bring about a new perspective to robot perception and simplify the autonomy pipeline. We demonstrate this through the design of a MEMS-driven LiDAR and perform compensation in two ways - (i) onboard IMU, and (ii) external feedback of robot pose at a high rate. 

\begin{figure}
    \centering
    \begin{subfigure}[b]{.34\linewidth}
        \centering
        \includegraphics[width=\linewidth]{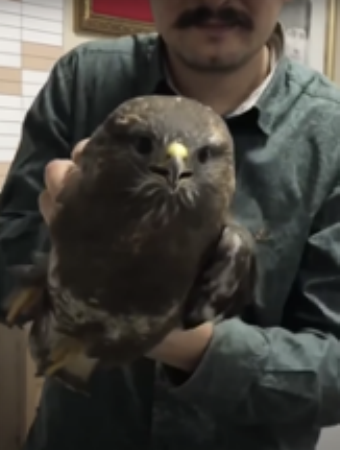}
        \caption{start of video}
    \end{subfigure}
    \begin{subfigure}[b]{.31\linewidth}
        \centering
        \includegraphics[width=\linewidth]{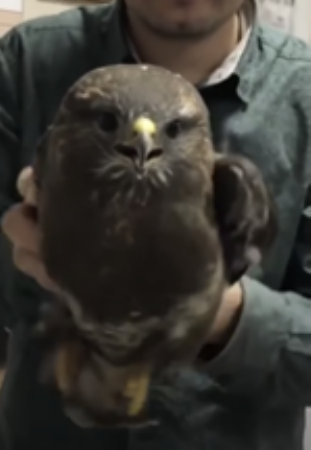}
        \caption{midway}
    \end{subfigure}
    \begin{subfigure}[b]{.27\linewidth}
        \centering
        \includegraphics[width=\linewidth]{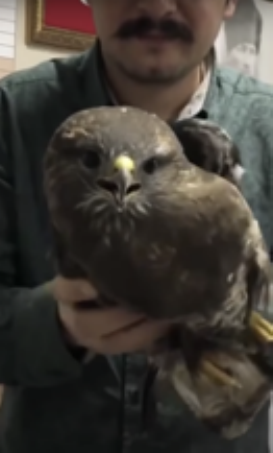}
        \caption{end}
    \end{subfigure}
    \begin{subfigure}[b]{.95\linewidth}
        \centering
        \includegraphics[width=\linewidth]{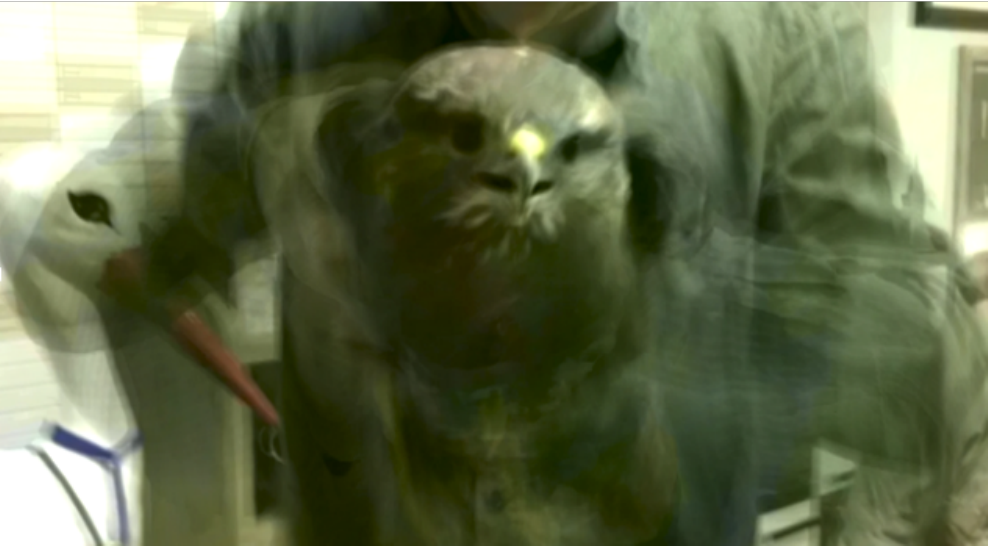}
        \caption{Average of all images in video}
    \end{subfigure}
    \caption{Biological motion compensation. The position and the angle of the head of the hawk remain stable despite body motion to provide the hawk an stabilized vision. 
  \href{https://www.youtube.com/watch?v=aqgewVCC0k0}{https://www.youtube.com/watch?v=aqgewVCC0k0}}
    \label{fig:hawk}
    \vspace{-15pt}
\end{figure}




We are inspired by animal eyes that have fast mechanical movements that compensate for motion, in real-time and at high accuracy \cite{alldieck2017optical}. In Fig.~\ref{fig:hawk}, we show frames $V(t)$ from a video of a hawk (\emph{Buteo jamaicensis}) being moved by a human trainer \cite{youtube_2020}. We also show the average of the video $\sum_t \frac{V(t)}{T}$ over a time interval $T$. Note that the averaged image shows motion blurring, except where the eagle mechanically compensates for the shifts. We envision biologically-inspired motion compensation that happens during sensing. These sensors need to adaptively change their orientation, in real-time, and in concert with robot goals such as mapping or navigation. Effectively, the rotation matrix $R$ must cancel out robot motion to provide a "stable" view of a scene. 


\subsection{MEMS Mirror-enabled Adaptive LIDAR}

The ability to reorient sensor pose could have many uses in robotics, particularly in image alignment during motion such as in SLAM. If the camera and robot are rigidly attached, then the camera experiences all the motion the robot experiences, including jitter and other potential disturbances that are detrimental to the Visual SLAM task. This could result in spurious features, errors in localization, and incorrect feature association leading to an inaccurate map. In this paper, we describe a sensor design that can perform image reorientation of a LiDAR in hardware without the need for any software processing for such compensation. 
Previously, pan-tilt-zoom (PTZ) cameras have attempted to address this problem. However, they use mechanical actuation which can react in ones of Hz making it not suitable for tasks such as egomotion compensation in real-time. This is evidenced by the limited use of PTZ cameras on robots - most robots just have sensors rigidly attached.

Our designs break through these past difficulties by exploiting recently available microelectromechanical (MEMS) and opto-mechanical components for changing camera parameters. Opto-MEMS components are famously fast (many kHz), and they allow the changing of the LiDAR projection offset orientation during robot motion, such that the view of LiDAR is effectively static. By changing LiDAR views two orders of magnitude (or more) faster than robot motion, we can effectively allow for camera views to be independent of the robot view. In this work, we can compensate the LiDAR point cloud using an onboard IMU or external feedback such as motion tracking setup. More generally, such compensation allows the robot to focus on the control task while the camera can perform perception (which is required for the control task) independently, and greatly simplifies robot planning as the planner does not need to account for perception and just needs to reason about the control task at hand.

MEMS LiDAR optics have the advantages of small size and low power consumption~\cite{tasneem2020adaptive, kasturi2016uav, kimoto2014development}. 
Our algorithmic and system design contributions beyond this are:

\begin{itemize}
    \item We present the design of a novel LiDAR sensor adopting a MEMS mirror similar to this LiDAR MEMS scanner \cite{wang2020low}. This design enables wide non-resonant scanning angles for arbitrary orientations. We integrate this with two types of feedback (IMU and external sensors) to demonstrate quick and high-rate motion compensation. Fig.~\ref{fig:futuristic} shows the design of our sensor. 

    \item We describe and geometrically characterize our sensor, showing that compensation in hardware can reduce the number of unknowns for proprioceptive and exteroceptive tasks. In a simulation, we characterize the effect of compensation delay and compensation rate to identify benefits for robot perception. The quantitative and qualitative results of these simulations are shown in Sec.~\ref{sec:benefits}.
    \item We present the compensation control algorithms for our LiDAR. We further characterize the performance of compensation control through experiments and simulations in Sec.~\ref{sec:novel_lidar_design}.
    \item We show UAV flight with a proof-of-concept hardware prototype combining external feedback with the MEMS mirror for egomotion compensation. We enable UAV flight by tethering the MEMS modulator to the other heavy necessary components, like the laser, photodetector, optics, the driver circuit, and the signal processing circuitry. The frequencies of the mirror modulation and IMU measurement are much higher than typical robot egomotion. Our prototype MEMS compensated scan system can perform such compensation in under 10 ms. Please see the accompanying video for proper visualization, and see Fig. \ref{fig:fvsc}.

    \item We provide an implementation of the sensor in the Gazebo simulator. Using this simulated sensor, we propose a framework to adapt modern LiDAR SLAM pipeline to incorporate motion compensation. We adapt a modern LiDAR SLAM pipeline LIO-SAM~\cite{shan2020lio} to incorporate motion compensation to use such a sensor and demonstrate the utility of such motion compensation. We have open-sourced the sensor implementation, the UAV simulation environment, as well as our LIO-SAM adaptations \footnote{\url{https://github.com/yuyangch/Motion_Compensated_LIO-SAM} } .

    
    
    
\end{itemize}




\section{Related Work}

\noindent \textbf{Small, compact LiDAR for small robotics:} MEMS mirrors have been studied to build compact LiDAR systems \cite{tasneem2020adaptive, kasturi2016uav, kimoto2014development}. For instance, Kasturi et al. demonstrated a UAV-borne LiDAR with an electrostatic MEMS mirror scanner that could fit into a small volume of 70 mm $\times$ 60 mm $\times$ 60 mm and weighed about only 50 g  \cite{kasturi2016uav}. Kimoto et al. developed a LiDAR with an electromagnetic resonant MEMS mirror for robotic vehicles \cite{kimoto2014development}. 

\noindent \textbf{Comparison to software-based compensation:} Motion compensation techniques and image stabilization techniques have been widely used in image captures. Similar to imaging devices, LiDAR point cloud shows point cloud blurring, motion artifacts caused by the motion of the LiDAR or the motion of the target object. Some software-based LiDAR motion compensation use ICP (iterative closest point) \cite{neuhaus2018mc2slam} and feature matching \cite{gojcic2019perfect} to find translation and rotation of successive point clouds. Software-based compensation for robotics motion has been studied in great detail in SLAM algorithms~\cite{taketomi2017visual} or expectation-maximization (EM) methods\cite{phelps2019blind}. Software-based motion compensation have a relative high computation barrier for micro-robotics and may degrade if the point cloud have large discrepancy. Some of the software-based motion compensation relies on the capture of a full frame of point cloud, so it cannot capture the motion frequency higher than the frame rate. For most of the LiDAR (other than flash LiDAR), especially the single scanning beam MEMS LiDAR, the rolling shutter effect caused by the LiDAR motion jitter remains a problem.  In contrast to these approaches, we wish to compensate the sensor in hardware, during image capture. Hardware LiDAR motion compensation has several benefits. First, the compensation can be implemented to every LiDAR scanning pulse (for 2D MEMS based LiDAR), which can correct the rolling shutter effect and improve the motion response range. Second, the motion compensation algorithm is very simple and can be implemented on a low-power microcontroller or FPGA. Third, even if the hardware motion compensation still have some errors, it provides a better initialization for the following software compensation.

These ideas are closer to how PTZ cameras track dynamic objects \cite{li2009tracking,hrabar2011ptz} and assist with background subtraction \cite{suhr2010background}. However, compared to these approaches, we can tackle egomotion of much higher frequencies, which is a unique challenge of micro-robots. We compensate signals much closer to those seen in adaptive optics and control for camera shake \cite{antonello2013imu, tyson1999performance, ben2004jitter}. In addition, our system is on a free moving robot, rather than a fixed viewpoint.

\noindent \textbf{Motorized gimbals:} Comparing to motorized image stabilization systems \cite{jia2017system}, MEMS mirrors not only have smaller size and lighter weight, but their frequency response bandwidth are better than the bulky and heavy camera stabilizer. The MEMS mirror response time can be less than 10 ms or even less than 1 ms. The servo motor of the camera stabilizer has a bandwidth width less than 30 Hz because they are bulky and have heavy load \cite{li2020nonorthogonal, sagitov2017towards}. This results in a response time higher than 10 ms. 

\noindent \textbf{Motion compensation in displays and robotics:} 
Motion-compensated MEMS mirror scanner has been applied for projection, \cite{gruger20073}, where hand-shake is an issue. In contrast, we deal with the vibration of much higher frequencies, and our approach is closest to adaptive optics for robotics. For example, \cite{taketomi2015zoom, taketomi2015focal} change the zoom and focal lengths of cameras for SLAM. Our prior work~\cite{tasneem2020adaptive} changed zoom in LiDARs. In this work, we compensate using small mirrors, utilizing a rich tradition of compensation in device characterization\cite{maksymova2019detection} and to improve SNR \cite{hayakawa2016gain}. Compared to all the previous methods, we are the first to show IMU-based LiDAR compensation with a MEMS mirror in hardware. 

\noindent \textbf{LiDAR SLAM}: Ever since the seminal work of ~\cite{zhang2014loam}, successive LIDAR SLAM designs largely follow a LiDAR SLAM architecture similar to Fig.~\ref{fig:rotating_lidar_slam_pipeline}, where the front end consists of De-skew and Feature extraction stages, while the back end usually consists of ICP and Pose Graph Optimization packages such as g2o~\cite{kummerle2011g} or GTSAM~\cite{dellaert2012factor} that globally optimizes the odometry information as estimated by LiDAR visual odometry. Successive efforts moved towards improvement in the following sub areas: 1) tightly coupling LiDAR and IMU ~\cite{ye2019tightly}; 2) updating the backend PGO optimizer~\cite{shan2020lio}; 3) updating the back end's ICP~\cite{yang2020teaser}; 4) updating the front end's point-cloud data structure to do away with ICP's feature dependence~\cite{xu2022fast}
\emph{Nevertheless, to the best of our knowledge, all existing LiDAR SLAM systems are designed for LiDARs that are rigidly attached, via fixed joints, to robots and vehicles.}

\noindent \textbf{Sensor reorientation in  Active SLAM:} There has been a lot of work in the area of perception-aware path planning. A basic assumption of this line of work is that the sensor is rigidly attached to the robot, and therefore, its field of view can be changed only by changing the pose of the robot. 
\cite{costante2016perception}\cite{sadat2014feature}\cite{deng2018feature} improve SLAM accuracy by actively changing the robot trajectory to improve the field-of-view. Our sensor can simplify these works by changing the FoV in hardware without requiring additional constraints on the path planning.

\begin{figure*} 
  \centering
  \includegraphics[width=.9\textwidth]{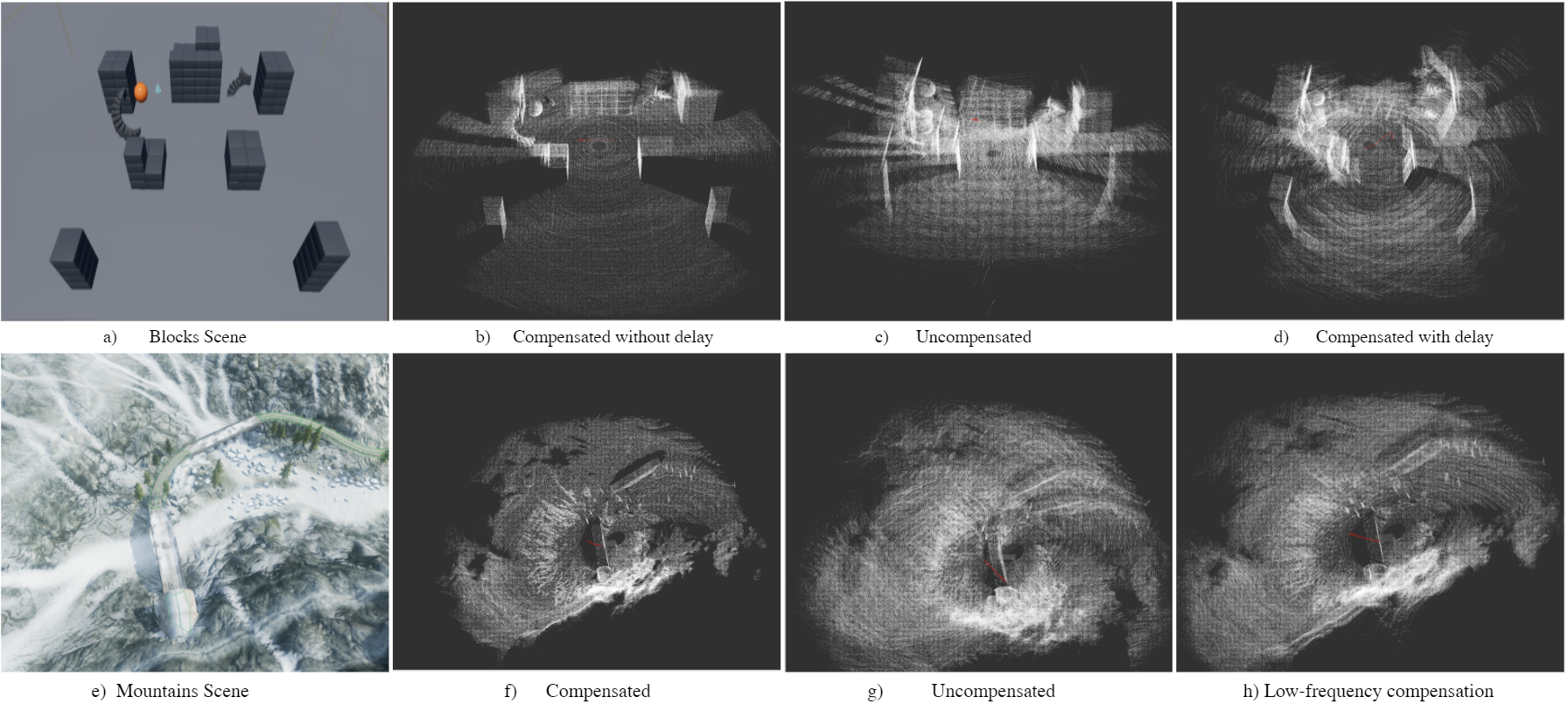}
  \caption{(a) Representative simulation scenario - Blocks scene (b) Mapping the Blocks scene with compensation at 55Hz and no delay (c) Mapping the Blocks scene without compensation  (d) Mapping the Blocks scene with compensation at  55Hz and delay of 150ms. (e) Mountains scene  (f) mountains scene simulation with 55Hz compensation and 0ms delay (g) mountains scene simulation without compensation. (h) mountains scene simulation with 5Hz compensation and 0ms delay.  }
  \label{fig:sim_qual}
  \vspace{-10pt}
\end{figure*}

\section{Understanding the Benefits of Compensated LiDAR in Simulation}
\label{sec:benefits}

\color{black}

\subsection{Basic LiDAR geometry}

A MEMS-based LIDAR scanning system consists of a laser beam reflected off a small mirror. Voltages control the mirror by physically tilting it to different angles. This allows for LIDAR depth measurements at the direction corresponding to the mirror position. Let the function controlling the azimuth be $\phi(V(t))$ and the function controlling the elevation be $\theta(V(t))$, where $V$ is the input voltage that varies with time step $t$. 

To characterize our sensor, we use the structure-from-motion (SFM) framework with the LIDAR projection matrix $\mathbf{P}$ and the robot's rotation $\mathbf{R}$ and translation $t$

\begin{equation}
    \mathbf{P} = 
    \begin{bmatrix}
      \mathbf{R} & t\\
      0 & 1
    \end{bmatrix}
\end{equation}


In our scenario, the `pixels' $\mathbf{x}$ relate to the mirror vector orientation $(\theta(V(t), \phi(V(t))$ on a plane at unit distance from the mirror along the z-axis, and are obtained by projections of 3D points $\mathbf{X}$. 
Many robotics applications need point cloud alignment across frames, which needs us to recover unknown rotation and translation that minimizes the following optimization. 

\begin{equation}
    \min_{\mathbf{R} , t} \| \mathbf{x} - \mathbf{P} \mathbf{X} \|.
\end{equation}

This optimization \emph{usually happens in software, after LiDAR and IMU measurements}~\cite{thrun2002probabilistic}. Our key idea is that the MEMS mirror provides an opportunity to compensate or control two aspects of the projection matrix $\mathbf{P}$ \emph{before capture, in hardware.} In this paper, we propose to control a new aspect of the SFM equation in hardware: the rotation matrix $\mathbf{R}$. 
Given the robot pose (from onboard IMU or other sensing) and the intrinsic matrix, we can easily perform post-capture translation estimation. 

\begin{equation}
    \min_{t} \| \mathbf{x} - \mathbf{P} \mathbf{X} \|. \label{simpl_opt}
\end{equation} 

In other words, hardware compensation with MEMS mirrors simplifies the post-capture LIDAR alignment methods to \emph{just finding translation $t$}, allowing for lightweight and low-latency algorithms to be used with minimal computational effort. 

\begin{figure*}
    \begin{center}
    \begin{subfigure}[b]{0.4\linewidth}
        \centering
        \includegraphics[width=\textwidth,trim={12pt 10 13 25},clip]{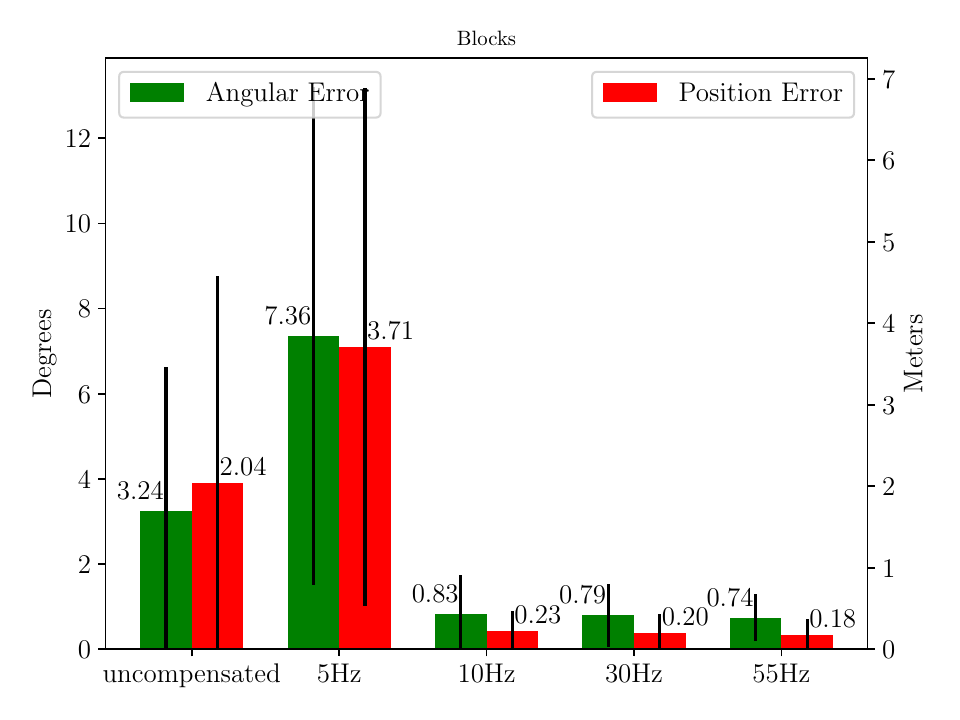}
        \caption{Odometry error vs compensation rate (Blocks scene)}
        \label{fig:sim_quant_blocks_comprate}
    \end{subfigure}
    \hspace{30pt}
    \begin{subfigure}[b]{0.4\linewidth}
        \centering 
        \includegraphics[width=\textwidth,trim={12pt 10 13 25},clip]{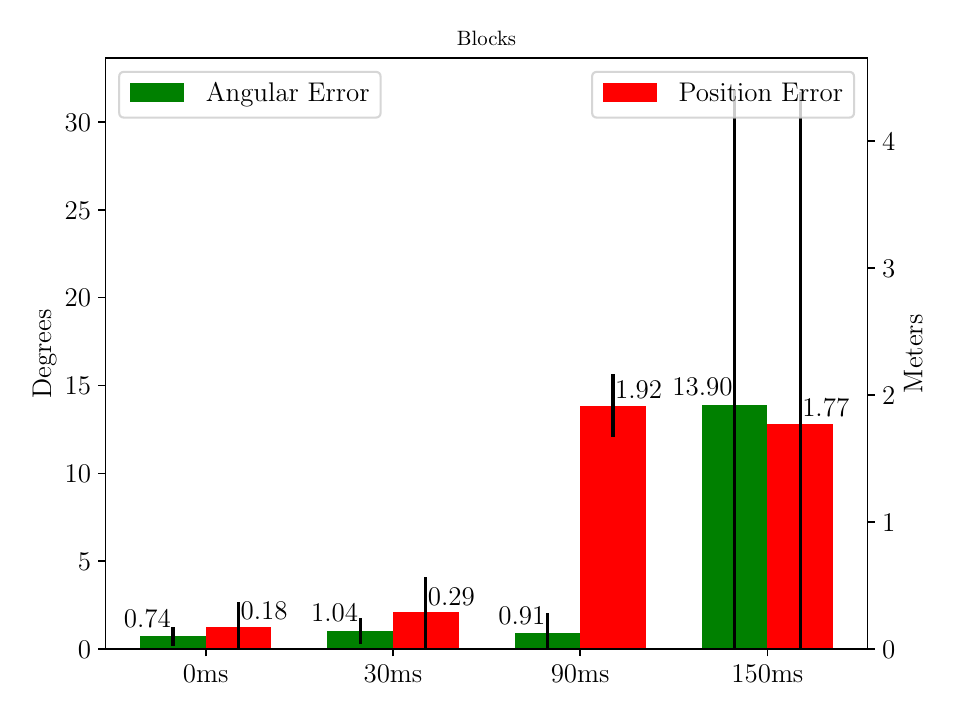}
        \caption{Odometry error vs compensation delay (Blocks scene). 55Hz Compensation}
        \label{fig:sim_quant_blocks_delay}
    \end{subfigure}
    \end{center}
    \begin{center}
    \begin{subfigure}[b]{0.4\linewidth}  
        \centering 
        \includegraphics[width=\textwidth,trim={12pt 10 13 25},clip]{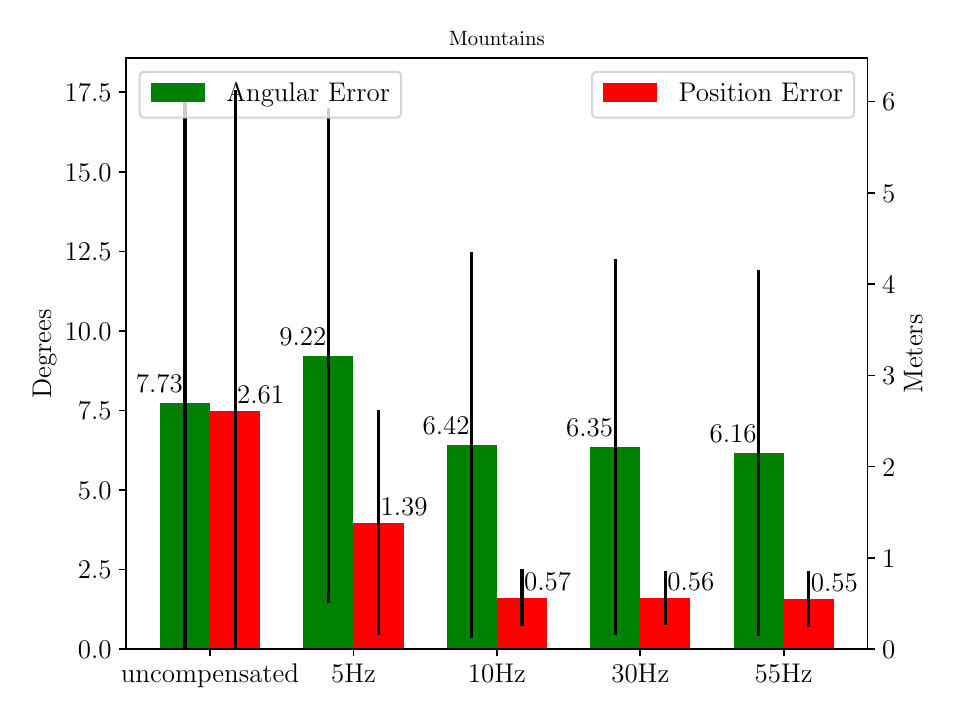}
        \caption{Odometry error vs compensation rate (Mountains scene)}
        \label{fig:sim_quant_mountains_comprate}
    \end{subfigure}
    \hspace{30pt}
    \begin{subfigure}[b]{0.4\linewidth}   
        \centering 
        \includegraphics[width=\textwidth,trim={12pt 10 13 25},clip]{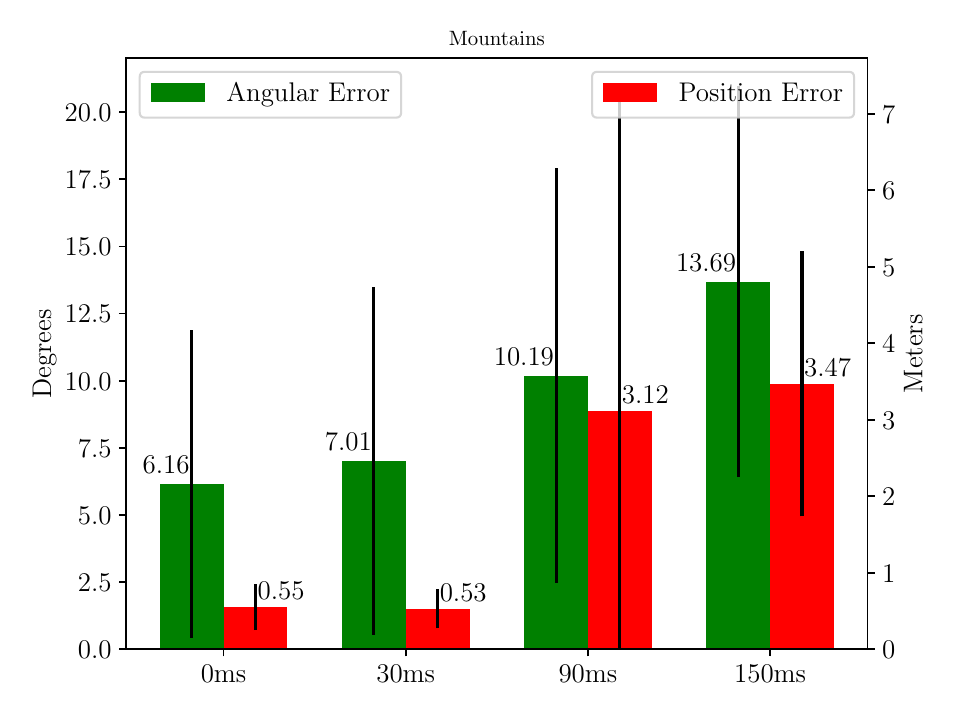}
        \caption{Odometry error vs compensation delay (Mountains scene). 55Hz Compensation}
        \label{fig:sim_quant_mountains_delay}
    \end{subfigure}
    \end{center}
    \caption{ UAV odometry error while varying compensation rate and compensation delay in two scenes}
    \vspace{-10pt}
    \label{fig:sim_quant}
\end{figure*}

\color{black}
\subsection{Benefits of IMU-compensated LiDAR in SLAM}
\label{sec:benefit of motion compensation in SLAM}
We demonstrate the benefits of motion compensated LiDAR in simulation. Our setup is as follows - we use Airsim \cite{shah2018airsim} running on Unreal Engine 4 for realistic perception and visualization. We tested two scenarios - a scene with geometric objects, called {\it Blocks scene} shown in Figure~\ref{fig:sim_qual}(a), and an outdoor scene with a bridge and mountains, called {\it Mountains scene} shown in Figure~\ref{fig:sim_qual}(e). In both scenes, the LiDAR is mounted on a prototype quadrotor UAV. We run LOAM~\cite{zhang2014loam}, an open-source state-of-the-art LiDAR SLAM system to map the environment and localize the UAV.

As described earlier, motion compensation can be achieved through various means such as a gimble, active compensation of a pan-tilt-zoom camera or MEMS-based hardware compensation like our system. The differences between these methods are along two dimensions - (i) latency of compensation, called {\it compensation delay} from now on, and (ii) number of times we can compensate in a second, called {\it compensation rate}. By varying these two parameters in simulation, we compare each method's performance. In order to systematically compensate based on IMU input, we perform some pre-processing of the IMU data. To smooth out the high angular velocity body movements, an angular moving average LiDAR stabilization algorithm is implemented. This method stores the past UAV orientations in a sliding, fixed length queue, and reorients the mounted LiDAR towards the average of the past orientations. The average of the orientations is calculated through Linear Interpolation (LERP) of the stored quaternions. We detailed our calculations in ~\ref{subsubsec:fov_stabilization}

The method is also known as Quaternion $L_2$-mean~\cite{hartley2013rotation}. Given the relative short duration of the sliding window, and the relatively small range of rotation that's covered during simulation flights, the prerequisite of using this method is met. It helps remove the impulsive jerky movements that may be observed by the LiDAR, akin to a low-pass filter.


In the experiment, the UAV performs three back-and-forth lateral flights between two way points. During the alternation of way points, the UAV reaches ~130$^\circ$/s in the X body axis. The mounted LiDAR is configured at 16 channels, 360 degree horizontal FoV, 30$^\circ$ vertical FoV and with 150,000 Hz sample rate, akin to commercially available LiDARs. 




To quantify performance, we calculate the{\it odometry error}, the difference between the ground truth UAV positions and those positions estimated by LOAM. Figure~\ref{fig:sim_quant} show the results from our simulations for Blocks scene and Mountains scene. We set the compensation rate to five different values - uncompensated, 5Hz, 10 Hz, 30 Hz, and 55 Hz. We set the compensation delay to five values - no delay (0 ms), 30 ms, 90 ms and 150 ms. 


Both the position error and angular error are high when the compensation rate is uncompensated or 5 Hz in the Blocks scene (Figure~\ref{fig:sim_quant_blocks_comprate}). It is significantly lower for 10 Hz, 30 Hz and 55 Hz. This shows that smaller rates of compensation as performed by a mechanical gimbal or a PTZ camera (which operate at 5 Hz or lower) are far less effective than a faster compensation mechanism such as the one proposed by us. Similarly, the error in position as well as orientation is low when the compensation delay is either 0ms or 30 ms (Figure~\ref{fig:sim_quant_blocks_delay}). 

For larger compensation delays such as 90ms and 150 ms, the error is several times that of when the compensation delay is 30 ms. This shows that as the compensation delay is higher, as it could be with software-based compensation on low-power embedded systems, it is far less effective and leads to greater error in trajectory estimation. This further argues for a system such as ours that is able to perform compensation in hardware, and therefore at a higher rate. The trends are similar, albeit less pronounced in the Mountains scene where features are much less distinct and feature matching is more challenging in general. This proof-of-concept set of simulations encouraged us to build our proposed system.

\section{Novel LiDAR Design}
\label{sec:novel_lidar_design}
We propose a simple and effective design, where the MEMS mirror and photodetector are placed on a movable head. For image stabilization, we are also able to place the IMU there. A LiDAR engine and accompanying electronics are tethered to this device, which can be light and small enough for micro-robots. To enable both the LiDAR scanning and compensated scanning at high rate, it is important to understand the characterization of the MEMS scanner. 
\subsection{The MEMS mirror}

All the compensation effects and size advantages described so far will be nullified if the MEMS mirror cannot survive the shock, vibration and shake associated with real-world robots. Here we analyze the robustness of the MEMS mirror device for such platforms. Most MEMS mirrors rely on high-quality factor resonant scanning to achieve wide field-of-view (FoV), which leads to heavy ringing effects and overshoot with sudden changes of direction \cite{milanovic2017closed, wang2020mems}. A suitable MEMS mirror for motion-compensated scanning is expected to have a wide non-resonant scanning angle, smooth and fast step responses, can operate under common robotics vibration and can survive shock. To achieve this goal, we adopt a popular electrothermal bimorph actuated MEMS mirror design \cite{jia2009electrothermal, wang2019large} to build this MEMS mirror. The employed MEMS mirror is fabricated with Al/SiO$_2$ based inverted-series-connected (ISC) bimorph actuation structure reported in \cite{jia2009electrothermal}. This type of MEMS mirror has the advantages of simple and mature fabrication process \cite{zhang2015fast, wang2017ultra}, wide non-resonant scanning angle, linear response and good stiffness.
A new electrothermal MEMS mirror is designed and fabricated with the adaption of the motion compensation application. We note that other previously reported MEMS mirrors with electrothermal actuators, electrostatic actuators, or electromagnetic actuators may also be applicable to the motion compensated LiDAR scanning \cite{kasturi2016uav,ito2013system,wang2020low}. 
\subsection{Compensation Algorithm}
\label{sec:comp_algo}
In the previous sections, we saw the advantages of MEMS mirror-based compensation and the feasibility for use in a robotic LiDAR. Here we focus on the details of the hardware-based rotation compensation algorithm using MEMS mirror scanning LiDAR and sensing for the compensation. 

The MEMS mirror reflect a single ray of light towards a point in the spherical coordinate $\{\alpha,\beta,r\}$. The $\{\alpha,\beta\}$ are the two angular control input to the mirror to achieve such target.   We will first establish the local(robot) and global (world) frames, then introduce known helper conversion from spherical to Cartesian coordinates, and finally gets into the details of compensation.

\subsubsection{Preliminaries}
\paragraph {Coordinate System}
\label{subsubsec:coordinate-system}

Our LiDAR can compensate for rotation, but it can not compensate for translation. So all discussion heres on in drops the translation from $SE(3)$ and will only be focus on $SO(3)$.
Let the robot have rotation $\boldsymbol{R}_{robot}^w \in SO(3)$ relative to the world frame. In here, the frame of the un-moving base of the LiDAR sensor have Identity rotation $\boldsymbol{R}_{base}^w \in SO(3)$ and therefore identical $SO(3)$ transformation as the robot frame. 

\paragraph{Spherical-to-Cartesian Conversions}
It is important to outline the conversion from the spherical, which is the control coordinate, to normal Cartesian coordinate. 
Points in the spherical coordinate $\{\alpha,\beta,r\}$ can be converted to Cartesian coordinate via known equations,

\begin{equation}
   p_{cartesian} =   
   \begin{bmatrix}
    x \\
    y \\
    z \\
    \end{bmatrix} 
    =
    \begin{bmatrix}
    r\cos{\alpha} \cos{\beta} \\
    r\cos{\alpha} \sin{\beta} \\
    r\sin{\alpha}\
    \end{bmatrix}
    \label{eqn:sphere-to-cartesian}
\end{equation}

and vice versa:

\begin{equation}
   p_{spherical} =   
   \begin{bmatrix}
    \alpha \\
    \beta \\
    r
    \end{bmatrix} 
    =
    \begin{bmatrix}
    \arctan{\frac{z}{\sqrt{x^{2} + y^{2}}}} \\
    \arctan{\frac{y}{x}} \\
    \sqrt{x^2+y^2+z^2}
    \end{bmatrix} 
    \label{eqn:cartesian-to-sphere}
\end{equation}

Note that both $p_{cartesian}$ and $p_{spherical}$ are points located in the robot's local coordinate frame, $\boldsymbol{R}_{robot}^w$. Other literature's refer to this frame as the local frame, or camera frame.

\paragraph{Spatial Scanning}
\begin{figure}[h]
    \centering
    \begin{subfigure}[t]{0.5\textwidth}
    \includegraphics[width=0.9\linewidth]{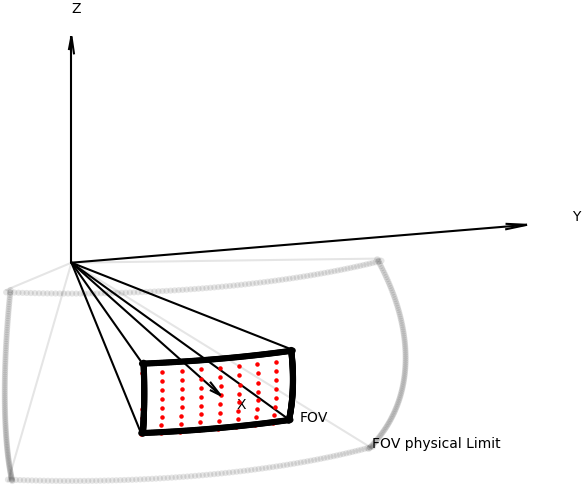}
    \caption{Depicting Spatial Scanning Grid in the sensor's base frame. In the real sensor, the resolution of the scanning grid is higher at 20x20. The input rotation here is zero. In other words, $\boldsymbol{R}_{control}=\boldsymbol{I}$.}
    \label{fig:spatial_scanning}
    \end{subfigure}%

    \begin{subfigure}[t]{0.5\textwidth}
    \includegraphics[width=0.9\linewidth]{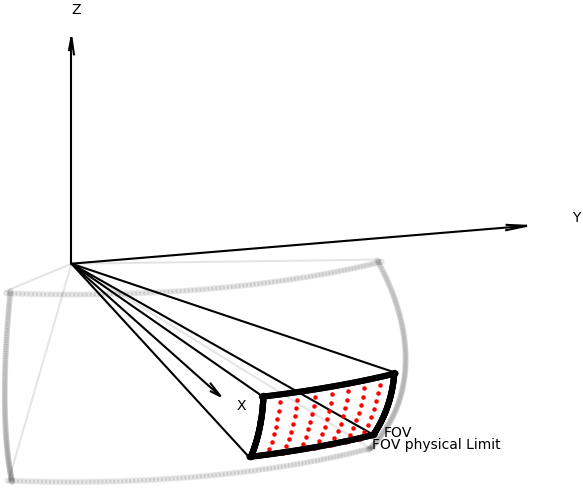}
    \caption{Depicting Spatial Scanning Grid in the sensor's base frame, The input rotation is none-zero here.}
    \label{fig:spatial_scanning_none_0}
    \end{subfigure}%
    \caption{}
    
\end{figure}

A set of $i$ spherical control coordinates $\{\alpha_i,\beta_i,r_i\}$  defines the scanning pattern of the LiDAR. We use $r_i=1$ for unit length vectors. In our setup, $\{\alpha_i,\beta_i\}$ defines a rectangular scanning grid in the spherical coordinate, whose center is the principle axis.  See figure~\ref{fig:spatial_scanning}. Within this limit, the mirror can direct its beam to any point desired by the user. 
\paragraph{Desired Sensor World Frame Rotation}
In our design, users can define a desired world frame rotation of the sensor, separately from the world frame rotation of the robot. The rotational decoupling of a sensor and a robot provides many benefits which we demonstrate through various application in this work.   Let $\boldsymbol{R}_{desired}^w \in SO(3)$ be the desired rotation target in the world frame. $\boldsymbol{R}_{desired}^w$ can be decided by the users. For example, it can be a slower changing rotation, relative to the robot's body frame. We demonstrated the benefit of such application in ~\ref{sec:benefit of motion compensation in SLAM}. We will touch on the exact details later in ~\ref{subsubsec:fov_stabilization}.  Other possibility includes aiming towards a specific world frame target $t \in \mathbb{R}^3$ which we will touch on later, in ~\ref{subsubsec:aiming}.

\subsubsection{General Rotation Compensation}
\label{subsubsec:general_rotation_compensation}
Given robot world frame rotation $\boldsymbol{R}_{robot}^w$, user desired sensor rotation $\boldsymbol{R}_{desired}^w$ and a set of spatial scanning, spherical, sensor input coordinates $\{\alpha_i,\beta_i,r_i\}$, we need to find the adjusted sensor input coordinates $\{\alpha^*_i,\beta^*_i,r^*_i\}$, in order to achieve user desired sensor rotation $\boldsymbol{R}_{desired}^w$. We will outline the calculations step by step.

\paragraph{Step1}
we first translate each $\{\alpha_i,\beta_i,r_i\}$ to Cartesian $p_{cartesian}$ by Eq.~\ref{eqn:sphere-to-cartesian}. This step is necessary, in order to calculate points transformation with rotation matrices.

\paragraph{Step2}
\label{paragraph:r_control}
The control rotation input to the sensor $\boldsymbol{R}_{control}$, is the difference between the desired world frame sensor rotation $\boldsymbol{R}_{desired}^w$ and the robot's current world frame rotation $\boldsymbol{R}_{robot}^w$.

Put it formally, Let $\boldsymbol{R}_{control}$ be the rotation from robot rotation $\boldsymbol{R}_{robot}^w$ to the desired rotation $\boldsymbol{R}_{desired}^w$ , therefore $\boldsymbol{R}_{desired}^w=\boldsymbol{R}_{control}\boldsymbol{R}_{robot}^w$. We have,
\begin{equation}
   \boldsymbol{R}_{control} =  \boldsymbol{R}_{desired}^w (\boldsymbol{R}_{robot}^w)^T
\end{equation}
Intuitively, When there is no difference between the desired sensor rotation and the robot rotation, where $\boldsymbol{R}_{desired}^w=\boldsymbol{R}_{robot}^w$ then $\boldsymbol{R}_{control}=\boldsymbol{R}_{desired}^w (\boldsymbol{R}_{desired}^w)^T=\boldsymbol{I}$. And $\{\alpha_i,\beta_i,r_i\}=\{\alpha^*_i,\beta^*_i,r^*_i\}$ This default orientation is shown in figure~\ref{fig:spatial_scanning}. When there is a difference however, an example is shown in Figure~\ref{fig:spatial_scanning_none_0}.

\paragraph{Step3}
Now, all points in the spatial scanning pattern $p_{cartesian}=\{x_i,y_i,z_i\}$ of the robot frame $\boldsymbol{R}_{robot}^w$ can be transformed to have the desired sensor world frame rotation $\boldsymbol{R}_{desired}^w$,:
\begin{equation}
   \label{eqn:r_control_transform}
   p_{desired-cartesian} =  \boldsymbol{R}_{control}p_{cartesian}
\end{equation}

substituting $\boldsymbol{R}_{control}$, we have

\begin{equation}
   p_{desired-cartesian} =  \boldsymbol{R}_{desired}^w (\boldsymbol{R}_{robot}^w)^Tp_{cartesian}
\end{equation}

\paragraph{Step4}
Finally, we can translate the rotated points $p_{desired-cartesian_i}$ back to the spherical coordinate $p_{desired-spherical_i}$, via Eq.~\ref{eqn:cartesian-to-sphere} for point $i$'s rotation control input to the sensor. Now that we have come to our answer for $\{\alpha^*_i,\beta^*_i,r^*_i\}$.

It is important to note that, this full $SO(3)$ compensation is only achievable because our LiDAR project individual point $p_i$ independently from other points in the set. In the case of a traditional camera or a commercially available LiDAR like Velodyne, The entire set of $p_i$ can be viewed as being projected as a group and correlate to each other. In these other sensors, Full $SO(3)$ compensation is not achievable, even if the sensors are mounted to the robot by a universal joint with 2 degree-of-freedoms $\alpha,\beta$. But we will also analyze this special case of grouped points re-projection since our LiDAR can achieve this 2-axis-only compensation.

\subsubsection{Special Case: 2-axes only compensation}
\label{subsubsec:special_case}
It is important to analyze the case where the sensor can only rotate in two axes relative to the robot. Such setup is commonly seen in robots with cameras mounted by a 2-axis gimbal, as well as PTZ cameras. Another applicable scenario is when we mount a commercially available Velodyne on a UAV via a universal joint, to perform LiDAR SLAM studies such as in ~\ref{sec:benefit of motion compensation in SLAM}  and design motion-compensated LiDAR SLAM ~\ref{sec:rotation_compensated_lidar_slam}. Furthermore, when it comes to the target aiming ~\ref{subsubsec:aiming}, 2-axis rotation is often preferred. Our sensor can perform such compensation as well.

In this subsection, we will outline the control not only for our sensor but all sensors, that mount on robots via joints or gimbals with 2-axis orientation controls.

In \ref{paragraph:r_control} we define $\boldsymbol{R}_{control}$ as the difference between the sensor orientation and the robot's orientation.  Since $\boldsymbol{R}_{control} \in SO(3)$ it requires at least 3-axis rotation control to achieve. 

We can collapse this $\boldsymbol{R}_{control}$ matrix into a rotation matrix that is the composition of two Euler angles. The new rotation matrix will not be identical to $\boldsymbol{R}_{control}$ , but it keeps the same sensor principle axis ray direction.  We herein refer to the collapsed version as $\boldsymbol{R}_{control}^*$.

Let $\boldsymbol{R}_{control}$ be limited to 2-axes rotation only:
    \begin{gather}
    \label{eq:special-case-rotation}
\boldsymbol{R}_{control}^*
 =
  \begin{bmatrix}
   \cos{\beta} & -\sin{\beta}&0 \\
    \sin{\beta} & \cos{\beta} & 0 &\\
   0 & 0 & 1
   \end{bmatrix}
  \begin{bmatrix}
   \cos{\alpha} & 0 & \sin{\alpha} \\
    0 & 1 & 0 &\\
   -\sin{\alpha} & 0 & \cos{\alpha} 
   \end{bmatrix}
\end{gather}

Here is how to find $\boldsymbol{R}_{control}^*$ from a given $\boldsymbol{R}_{control}$, step by step.

\paragraph{Step1}
Rotate the principle axis $\boldsymbol{e}_1$, with $\boldsymbol{R}_{control}$  in our case $\boldsymbol{e}_1=\{x=1,y=0,z=0\}^T$.
\begin{equation}
   \boldsymbol{e}_{rotated} =  \boldsymbol{R}_{control}\boldsymbol{e}_1
\end{equation}
$\boldsymbol{e}_{rotated}$ is now the new principle axis that our sensor should target.
Note that, $\boldsymbol{e}_{rotated}$ is closely related to the ray vector from robot to target in the aiming application, more on this later at ~\ref{subsubsec:aiming}.

\paragraph{Step2}

We then translate this Cartesian coordinate $\boldsymbol{e}_{rotated}$ vector into the Spherical coordinate, using Eq.~\ref{eqn:cartesian-to-sphere}. We will get $\{\alpha,\beta,1\}$. 
\paragraph{Step3}
Finally, we can use Eq. ~\ref{eq:special-case-rotation} to find the collapsed $\boldsymbol{R}_{control}^*$ with $\alpha,\beta$.

Our LiDAR can then use $\boldsymbol{R}_{control}^*$ to perform 2-axis only compensation. We can simply follow the same steps in ~\ref{subsubsec:general_rotation_compensation}, except we replaces $\boldsymbol{R}_{control}$ in Eq.~\ref{eqn:r_control_transform} with $\boldsymbol{R}_{control}^*$.

Further, this compensation can be readily extended to commercially available cameras and LiDARs (such as Velodyne) mounted on a universal joint to the robot frame or a 2-axis gimbal-mounted camera. The 2-axis angles $\alpha,\beta$ are enough to describe the two joint rotations.

\subsubsection{Rotational FoV Stabilization}
\label{subsubsec:fov_stabilization}
It is often desirable to have relatively slow rotating sensor world frame FoV in many SLAM related applications. We have demonstrated such benefit in ~\ref{sec:benefit of motion compensation in SLAM}. In here we go into details of how it is achieved with our sensor.

\paragraph{Quaternion $L_2-mean$}

Supposedly $q_1.....q_n$ is the world frame quaternions store in a queue data structure, representing the robot's world frame rotation in the last $n$ time stamps. We can find its average via LERP, summing and normalizing the quaternions as 4-vectors~\cite{hartley2013rotation}:
\begin{gather}
    q_{avg}
     = \frac{\sum_{i=1}^n q_i}{||\sum_{i=1}^n q_i||_2}
     \label{eqn:LERP}
\end{gather}

$q_{avg}$ can be converted into a rotation matrix $\boldsymbol{R}_{desired}^w$. Along with the robot's current world frame rotation $\boldsymbol{R}_{robot}^w$, we can find the adjusted spherical coordinate control input to our sensors $\{\alpha^*_i,\beta^*_i,r^*_i\}$, according to ~\ref{subsubsec:general_rotation_compensation}.

\subsubsection{Target Aiming}
\label{subsubsec:aiming}
Let $t_{target}^w \in \mathbb{R}^3$ be the target of interest in the world frame, and let $t_{robot}^w$ be the robot's current world frame translation. Then
 \begin{gather}
p_{aim}=(\boldsymbol{R}_{robot}^w)^{T}(t_{target}^w-t_{robot}^w)
\end{gather}
outlines the ray direction which we want to align our "principle axis", or the projection center point towards.
Following a very similar process as to ~\ref{subsubsec:special_case} we can find the controls:
\paragraph{Step1}

We can simply translate Cartesian coordinate $p_{aim}$ to spherical coordinate via Eq.~\ref{eqn:cartesian-to-sphere} to find $\alpha,\beta$.

\paragraph{Step2}
Then compose a $\boldsymbol{R}_{control}^*$ via Eq.~\ref{eq:special-case-rotation} for the entire scanning grid. We can simply follow the same steps in ~\ref{subsubsec:general_rotation_compensation}, except we replaces $\boldsymbol{R}_{control}$ in Eq.~\ref{eqn:r_control_transform} with $\boldsymbol{R}_{control}^*$

For all other sensors mounted on 2-axis gimbals or universal joints, $\alpha,\beta$ is enough to describe the joint inputs.

\subsubsection{MEMS Related details}
MEMS-related details, relating to the 1-dimension controls of each actuation axis $\{\alpha,\beta\}$, ~including analysis of
robot motion shock on the MEMS as well as preliminary pointcloud stitching, are included in the appendix.

\color{black}

\color{black}
\subsection{LiDAR Hardware Specifics} 
\begin{figure*}[!h]
  \centering
  \includegraphics[width=0.8\textwidth]{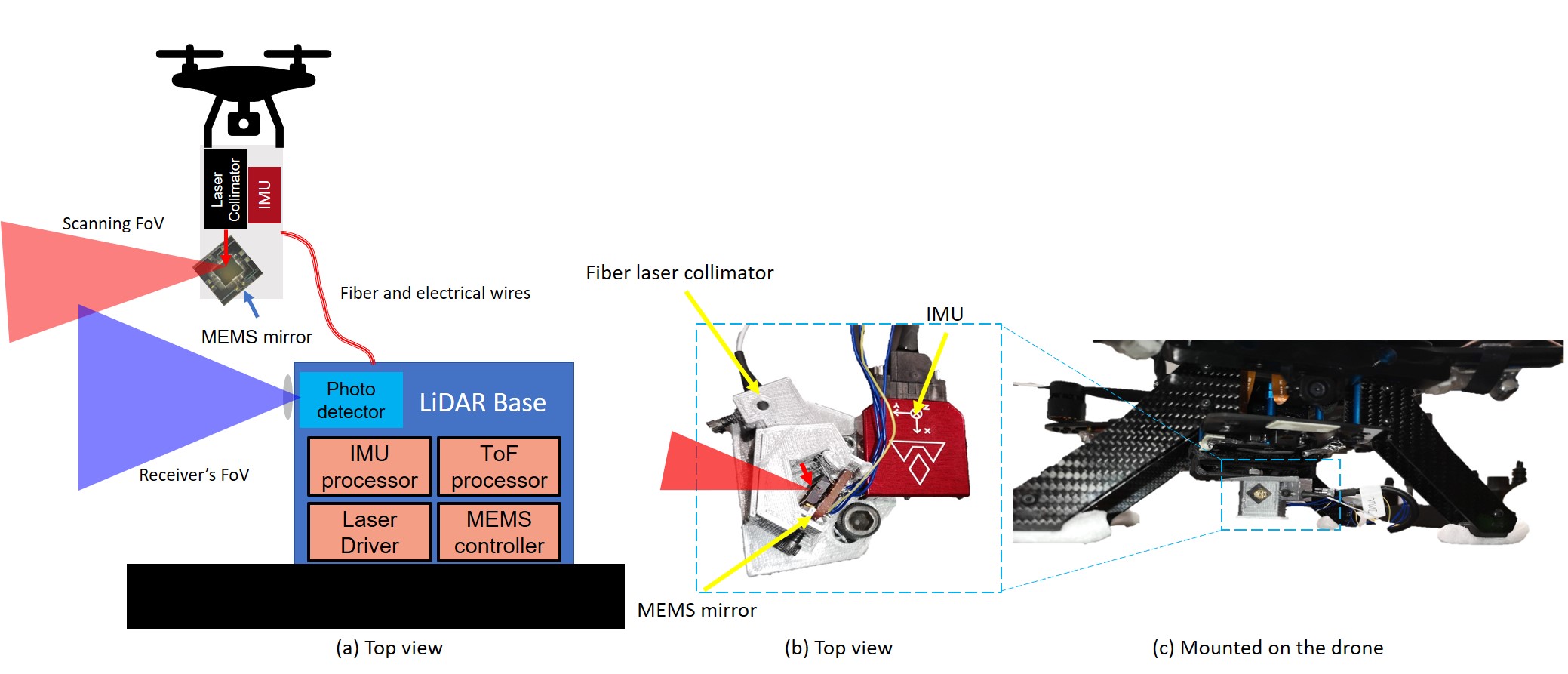}
  \caption{The movable LiDAR MEMS scanner head, which include the MEMS mirror, an IMU and a fiber laser collimator. (a) shows the top view and (b) shows the LiDAR scanner head mounted to the bottom of the UAV.}
  \label{fig:setup}
\end{figure*}

\begin{figure}[h]
    \centering
    \includegraphics[width=0.9\linewidth]{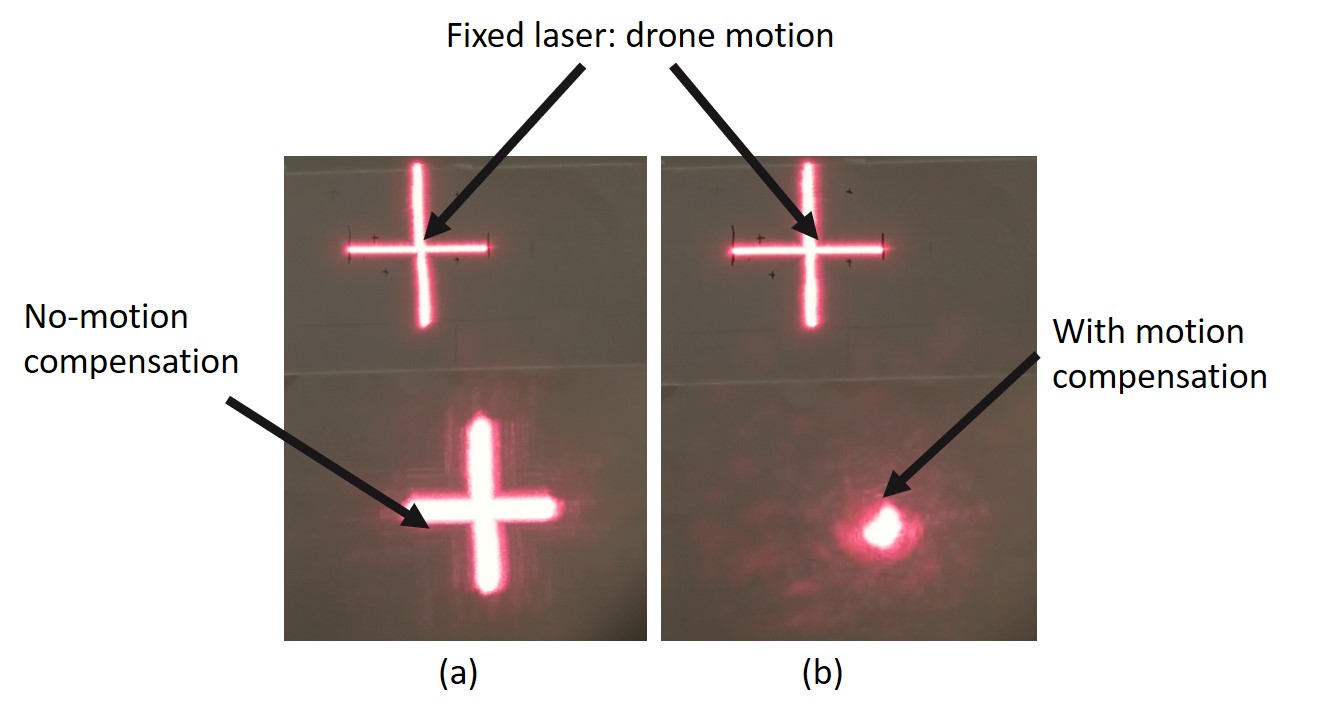}
    \caption{We use a visible laser to  compare the effect of motion compensation of our sensor. The upper laser trace indicates UAV motion, and the lower laser trace indicates the compensated/uncompensated scanning laser reflected from the MEMS mirror. The compensated MEMS scanning (right) shows a much smaller laser trace area than the uncompensated MEMS scanning result.}
    \label{fig:vis}
\end{figure}
\begin{figure*}[!ht]
  \centering
  \includegraphics[width=0.9\textwidth]{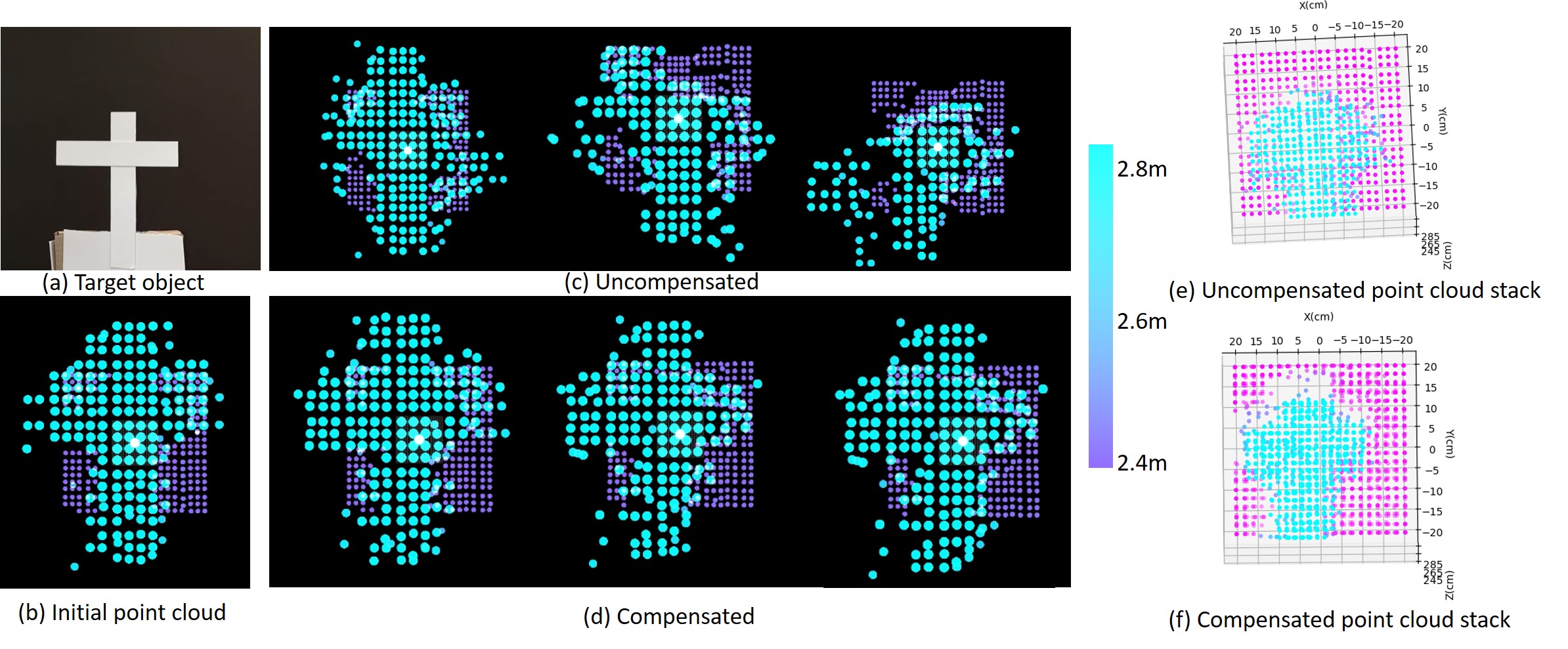}
  \caption{ Motion compensated LiDAR point cloud result with hand held motion disturbance. (a)The target object "+" placed 2.4 m from the LiDAR, along with (b), its initial point cloud scan. (c) and (d) show uncompensated vs. compensated scanning. The hand held rotation angle in (x, y) axes are $(-0.2 ^\circ, +1.4 ^\circ), (+1.0 ^\circ, +1.4 ^\circ), (-1.5 ^\circ, +1.7 ^\circ)$ and compensated angular shake range was $(-1.1 ^\circ, +1.4 ^\circ), (+1.2 ^\circ, -0.5 ^\circ), (-2.3 ^\circ, +0.5 ^\circ)$. (Please see supplementary video). (d and e) The stacking of 5 frame of point cloud of compensated and uncompensated results.}
    \label{fig:exp2}
\end{figure*}

Our prototype~(Fig.~\ref{fig:setup}) uses an InGaAs avalanche photodiode module (Thorlabs, APD130C). A fiber with a length of 3 m delivers the laser from the laser source to the scanner head. The gain-switch laser (Leishen, LEP-1550-600) is collimated and reflected by the MEMS mirror. The X-axis of the IMU (VectorNav, VN-100) is parallel to the neutral scanning direction of the MEMS mirror. The in-run bias stability of the gyroscope is $5-7^\circ$/hr typ. The scanner head sits on a tripod so that it can be rotated in the yaw and pitch directions. In the LiDAR base, an Arduino microcontroller is used to process the time-of-flight (ToF) signals, sample the IMU signals and control the MEMS mirror scanning direction. The data are sent to a PC for post-processing and visualization.

Since our motivation was to use micro-robots, our maximum detection distance is 4 m with a $80\%$ albedo object and the minimal resolvable distance is 5 cm. The maximum ToF measurement rate is 400 points/sec. According to the compensation algorithm described in the previous section, the MEMS mirror scanning direction is updated and compensated for motion at 400 Hz. We now describe our experiments. Please see the accompanying video for further clarification. 

\subsection{Compensation experiments with zero translation}
\subsubsection{Handheld Experiments}
 To demonstrate the effect of compensation, a visible laser is used instead of the LiDAR IR light to visualize the effect of tracking. We mount the LiDAR MEMS scanner on the UAV, as shown in Fig.~\ref{fig:setup}. The MEMS mirror desired scanning angle is set to a single point on the target object ($0^\circ$ by $0^\circ$) to make it easier for comparing. 
 
\color{black} Here the entire scanning grid $\{\alpha_i,\beta_i,1\}$ consist of one single point only at the projection center. We use the general compensation outline in ~\ref{subsubsec:general_rotation_compensation}
\color{black}
 
 The UAV together with the LiDAR scanner head is held with hand with random rotational motion in yaw/pitch direction. The upper laser trace comes from the laser rigidly connected to the UAV which indicates the UAV's motion. The lower trace is reflected from the MEMS mirror, which shows the compensated/uncompensated scanning laser. The results are shown in \autoref{fig:vis}. The MEMS scanning laser trace area of the compensated scanning is significantly smaller than the uncompensated scanning trace under similar rotational motion disturbance. The videos of the real-time compensation results is available in the supplementary materials.

Then the IR pulse laser is connected to run the LiDAR. An object of interest (in the shape of a +) is placed 2.4 m away from the LiDAR and at the center of the field of view and the background is at 2.8 m, as shown in Fig.~\ref{fig:exp2}(a). The MEMS mirror performs a raster scanning pattern with an initial field of view of $-3.5^\circ$$\sim$$+3.5^\circ$ in both axes to leave the room for compensation. Each frame has 20 by 20 pixels, and the frame refresh rate is 1 fps. To mimic robot vibration, the tripod is rotated randomly in the directions of yaw (Z-axis) and pitch (Y-axis), and the point clouds are shown in Fig.~\ref{fig:exp2}(d). Despite the motion of the LiDAR head, the point clouds are quite stable. The differences among all of the point clouds are generally less than 2 pixel in either axis, caused by measurement noise.

Fig.~\ref{fig:exp2}(c) shows the point clouds without compensated scanning, where the relative positions of the target object in the point clouds keep changing. The target object may come out of the MEMS scanning FoV without compensation. With a continuous rotation of 1.5 Hz in the Y-axis, the same structure may appear in multiple positions in the same frame of the point cloud, as shown in the 3rd figure of Fig.~\ref{fig:exp2}(c). Multiple frames of point cloud are stocked together and shown in the last column of Fig.~\ref{fig:exp2}. The object can still be identified in the compensated point cloud (Fig.~\ref{fig:exp2}(f)), but becomes fuzzy caused by the motion jitter when not compensated (Fig.~\ref{fig:exp2}(e)). The videos of the real-time compensation point cloud results is available in the supplementary materials.

\subsubsection{Motorized input experiments}
\label{subsec:motor_experiments}

\begin{figure}[h]
    \centering
    \includegraphics[width=0.9\linewidth]{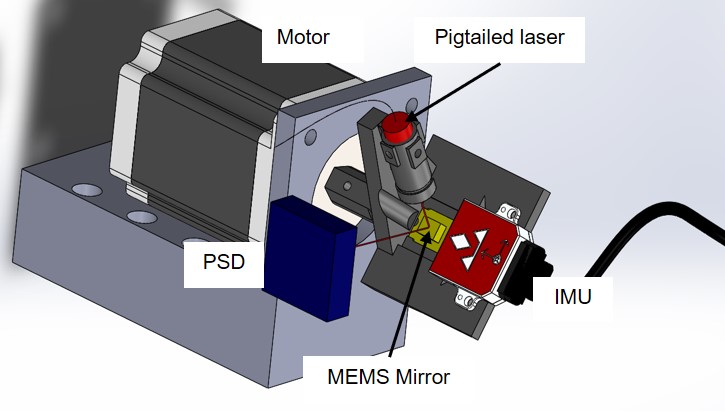}
    \caption{Input disturbance testing platform with stepper motor. }
    \label{fig:motor_PSD_platform}
\end{figure}

We use a separate platform to test the 1-d response of our mirror, with disturbance input from a stepper motor, refer to Fig.~\ref{fig:motor_PSD_platform}.~The MEMS mirror motion compensation system is controlled by an Arduino Mega. The IMU sends the data to the Arduino at 400 Hz. The data is processed, and the compensator $H_{g}(s)$ is implemented by the Arduino to get the MEMS angle and the desired driving voltages of the MEMS mirror. The two MEMS orthogonal scanning directions are assembled parallel to two IMU axes. To evaluate the compensation results, the reflected laser is captured by a PSD (position-sensitive detector) sensor fixed on the bench. The PSD sensor is placed 12 cm from the MEMS mirror. The PSD is for compensation evaluation only and is not in the controller loop. 


The motion-compensated MEMS scanner test is assembled on a step motor to test the compensation capability under various frequencies. The test bench, including the MEMS mirror, the IMU, and the pigtailed laser are fixed on the shaft of the stepper and rotate with the motor. One of the MEMS scanning directions is coincident with the motor rotation direction. The laser is delivered through a fiber. The stepper has a step size of $1.8^\circ$. With a micro-stepper controller, the approximate minimal step is as small as $0.018^\circ$ for smooth step translation control. The transient time of a $1.8^\circ$ step can be set from 30ms to 500ms. The motion compensation is tested in the pitch direction for smaller errors. The motor is placed horizontally to the ground.  



\begin{figure*}
    \centering
    \includegraphics[width=\textwidth]{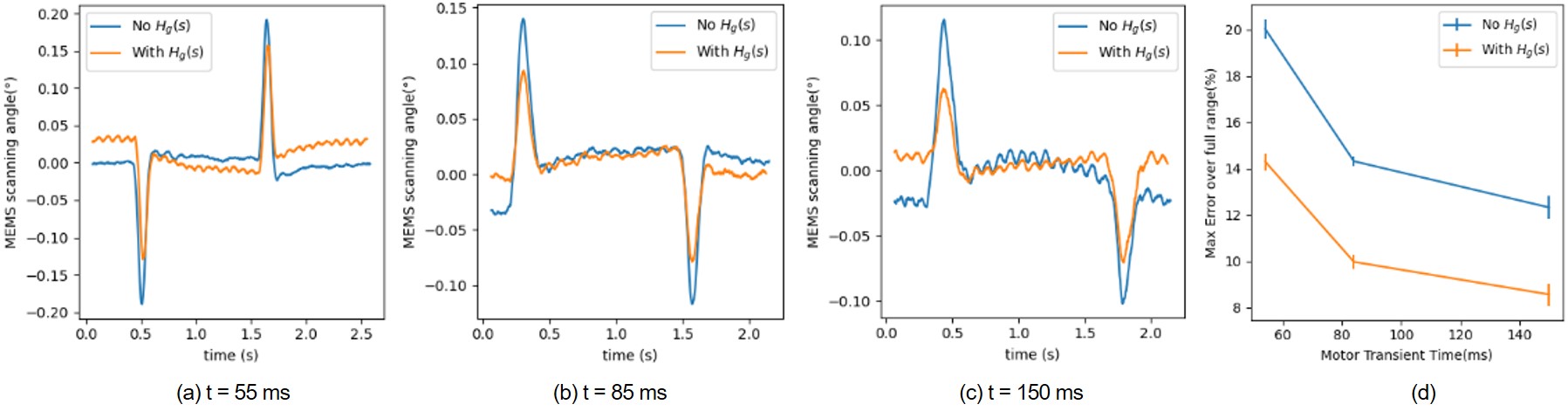}
    \caption{The motion compensation results comparison under different motor speed (t, motor transient time) and with and without the compensator $H_{g}(s)$. In these experiments, the mirror is commanded to aim at 0 degree, while the disturbance of 1.8 degrees is input to LiDAR base from a stepper motor. See Fig.~\ref{fig:motor_PSD_platform}. Therefore, ideally the MEMS scanning angle stays flat at 0 degree at all times. When the motor speed gets faster, the motion compensation errors become larger.  $H_{g}(s)$ can effectively improve the compensation scanning. }
    \label{fig:comp_vs_speed}
\end{figure*}

Fig. \ref{fig:comp_vs_speed} shows the motion compensation results comparison under various motor speed ($t$, motor transient time) and with and without the compensator $H_{g}(s)$. When the motor speed gets faster, the motion compensation errors increase, and $H_{g}(s)$ can effectively reduce the error. 

\begin{figure*}
    \centering
    \begin{subfigure}[b]{0.24\textwidth}
      \centering
      \includegraphics[width=\textwidth]{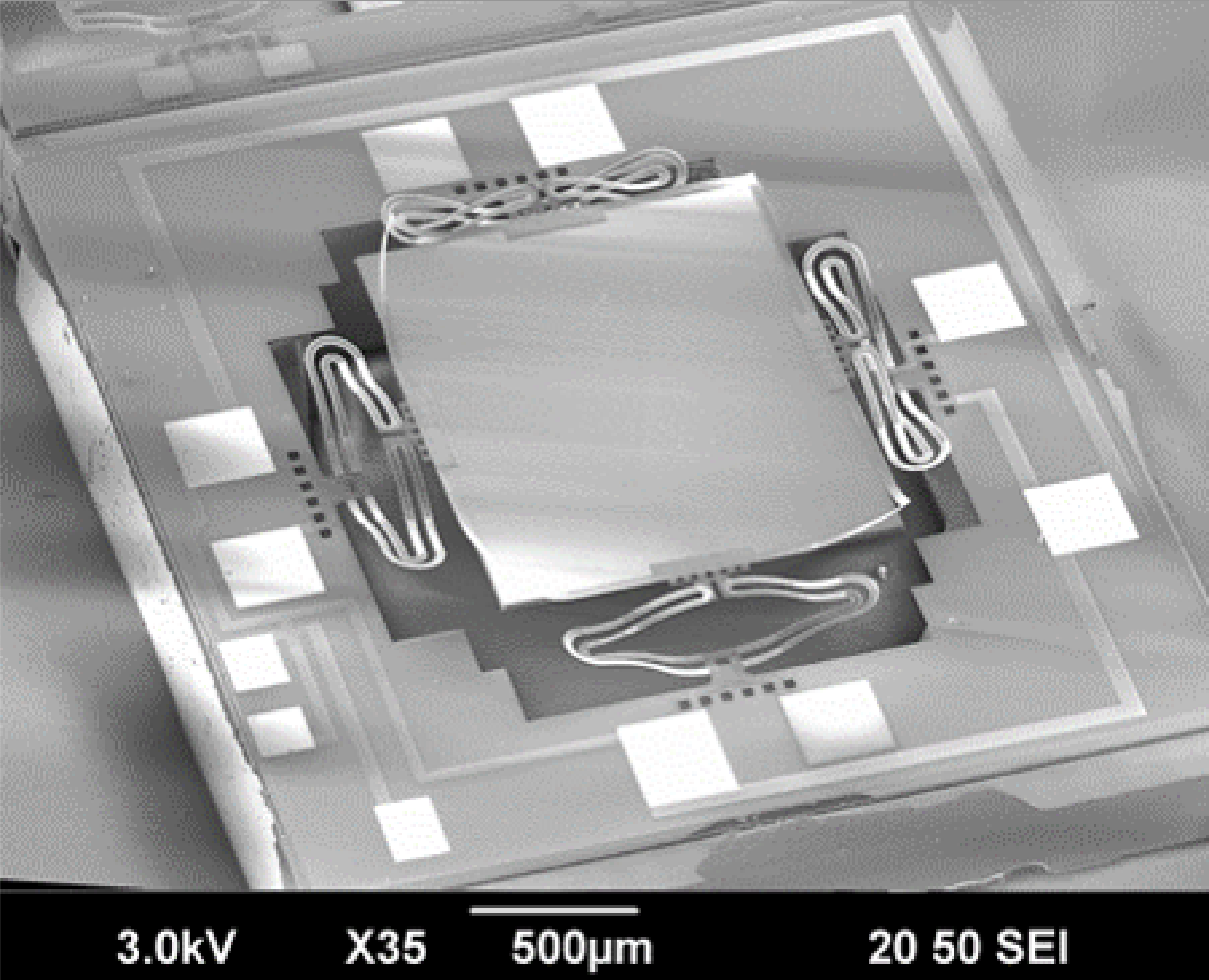}
      \caption{}
      \label{fig:memssem}
    \end{subfigure}%
    \hfill
    \begin{subfigure}[b]{0.24\textwidth}
      \centering
      \includegraphics[width=\linewidth]{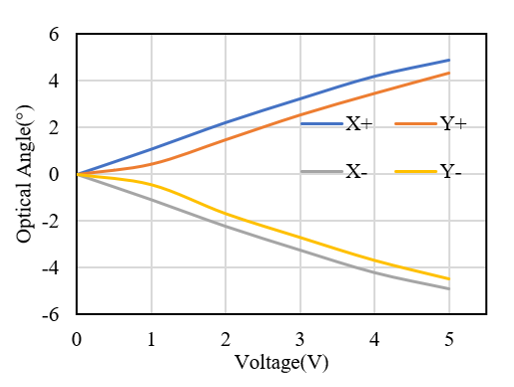}
      \caption{}
      \label{fig:memsangle}
    \end{subfigure}%
    \hfill
     \begin{subfigure}[b]{0.24\textwidth}
      \centering
      \includegraphics[width=\linewidth]{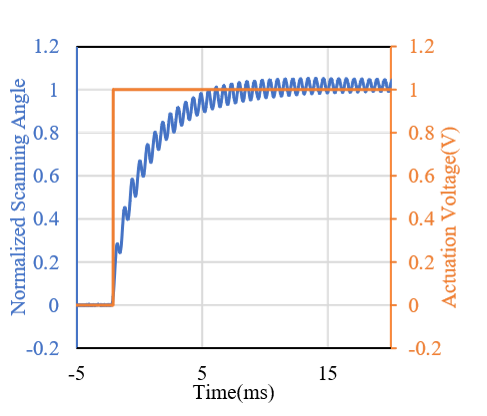}
      \caption{}
      \label{fig:memsstep}
    \end{subfigure}
    \begin{subfigure}[b]{0.24\textwidth}
      \centering
      \includegraphics[width=\linewidth]{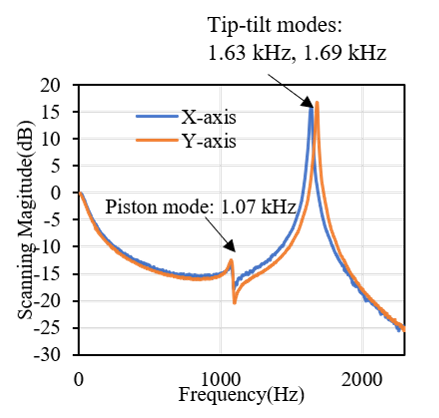}
      \caption{}
      \label{fig:memsfreq}
    \end{subfigure}
    \caption{(a) An SEM image of the fabricated MEMS mirror. (b) The optical scanning angle response of the MEMS mirror. (c) The step response of the MEMS mirror is 5 ms. (d)The frequency response of the MEMS mirror.}
    \label{fig:fvsc}
\end{figure*}


The motion compensation under continuous sinusoidal drive disturbance is also tested. The motor drives the MEMS mirror scanner head with sinusoidal motions. The MEMS compensated scanning system tries to compensate the scanning angle to an ideal direction. The effect with and without the compensator term is also compared, as shown in Table \ref{table:comp}.

\begin{table}
    \centering
    \caption{A comparison of the errors under step motion disturbance and continuous sinusoidal motion disturbance on a step motion with the $H_{g}(s)$ present or not.}
    \includegraphics[width=\linewidth]{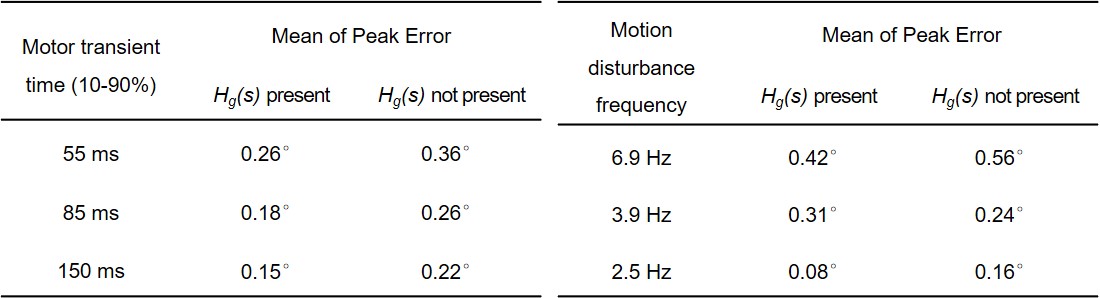}
    \label{table:comp}
\end{table}

\subsection{MEMS, Step Response and Robot Motion Shock}
We now describes the characteristics of MEMS mirror itself, particularly when it comes to disturbances from robot motion. see Fig.~\ref{fig:memssem}. In a later Sec.~\ref{sec:robotexp} we will show experiment results where the LiDAR is mounted on a UAV and perform target-aiming and spatial scanning tasks.

The mirror has a maximum actuation voltage of 5 V and a scanning FoV of $-4.8^\circ$ to $+5.2^\circ$ in the horizontal axis and $-3.8^\circ$ to $+4.3^\circ$ in the vertical axis (Fig. \ref{fig:memsangle}). The voltage to MEMS tilting angle response is approximately linear. The MEMS mirror can perform non-resonant arbitrary scanning or pointing according to the control signal. The cross-axis sensitivity is about 6 $\%$ in both axes. In the micro-controller, the voltage and the MEMS scanning angle are
approximated with a linear relationship with the cross-axis sensitivity
taken into consideration. The maximum error caused by the non-linearity is $0.3^\circ$.

The step response is 5 ms (Fig. \ref{fig:memsstep}(a)) with very small ringing. To test the frequency response, the frequency of the actuation voltage is swept and the actual tilt angle is measured by tracking the light beam reflected from the mirror plate using a position sensing detector (PSD), shown in Fig. \ref{fig:memsfreq}(b). The piston resonant mode is found at $f_1 $=1.07kHz and the tip-tilting resonant modes are at $f_2 = $1.63 kHz and  $f_3 =$ 1.69 kHz. 

The tip-tilt scanning response of the MEMS mirror is modeled with a 3rd order system according to \cite{li2017modelling}. The transfer function $H_1(s)$ can be expressed as,

\begin{equation} \label{H1s}
    H_1(s) = \frac{\frac{1}{\tau}\omega_n^2}{(s^2+2\omega_n\zeta s + \omega_n^2)(s+\frac{1}{\tau})}
\end{equation}

where $\tau$ is the thermal time constant, $\tau\approx t_r/2.2 = 2.3 ms$; $\omega_n$ is the natural resonant frequency of the mirror rotation, $\omega_n \approx 2\pi \text{(1.65 kHz)}$, and $\zeta$ is the damping ratio of the bimorph-mirror plate system, $\zeta \approx 1/2Q = 0.006$. Thus, the transfer function $H_1(s)$ of the MEMS mirror can be obtained by substituting and slightly tuning the parameters in Eq.~\ref{H1s}. 

Similar to \cite{wang2020low}, the MEMS mirror is actuated by the PWM signals with a voltage level of 0-5V. The PWM signal can be generated by an Arduino microcontroller at 15 kHz and 8 bit. The ringing of a step response is less than 2 $\%$ after about 10 ms. The minimal achievable step is 0.035$^\circ$ which is much smaller than the linearization error. 

We now show expressions for the acceleration and forces generated by a MEMS mirror scan. The small-angle tip-tilt scanning stiffness $k_{\mbox{r}}$ is

\begin{equation}
    k_{\text{r}} = I(2\pi f_{r})^2
\end{equation}

\noindent where $f_{r}$ is the resonant frequency of the tip-tilting modes ($f_2, f_3$); $I$ is the moment of inertia of the mirror plate alone its tip-tilting axis, 

\begin{equation}
    I = \frac{1}{12} m_{\text{plate}}(t^2+d^2)
\end{equation}

\noindent where $t$ is the thickness of the mirror plate, and $d$ is the length of the mirror plate. The rotation stiffness $k_{\mbox{r}}$ = 2.16e-6 N$\cdot$m/rad. With an external angular acceleration of $\ddot{\theta}$ alone on the mirror rotation axis, the excited mirror rotation $\theta$ is 

\begin{equation}
    \theta = -\frac{I\ddot{\theta}}{k_r} = \text{-1e-8} \cdot\ddot{\theta}.
\end{equation}

Take the mirror scanning step $0.25^\circ$ as the maximum tolerance of the excited mirror plate rotation, the tolerable external angular acceleration is $\ddot{\theta}$ = 44000 $\text{rad}$/$\text{s}^2$. The maximum angular acceleration of a commercialized robot is usually less than 1000 $\text{rad}$/$\text{s}^2$, and the excited MEMS mirror rotation is less than 6e-4$^\circ$ which can be ignored. Since this MEMS mirror has four identical actuators and the difference on the two axes alone are small, the excited MEMS mirror rotation under robot vibration can be ignored.

We now consider robot crash scenarios. The MEMS mirror can also survive most of the extreme vibration or mechanical shock without failure. The stiffness of the MEMS mirror under shock $k_{\mbox{p}}$ is:
\begin{equation}
    k_{\mbox{p}} = m_{\text{plate}} (2 \pi f_1)^2
\end{equation}

\noindent where $m_{plate}$ is the mass of the mirror plate. Thus, the stiffness the MEMS mirror in piston motion is $k_{p}$ = 3.2 N/m. The maximum allowable piston displacement of the mirror plate without failure is $d_{\mbox{max}}$= 200$\upmu$m. The maximum tolerable acceleration in the direction perpendicular to the mirror plate is $a_{max}$ is 

\begin{equation}
    a_{max} = \frac{k_{\text{p}} d_{\text{max}}}{m_{\text{plate}}} = 5500 \text{m}/\text{s}^2.
\end{equation}

For most commercial robots, maximum tolerable shock is under 1000 $\text{m}/\text{s}^2$. \emph{So the MEMS mirror can survive most of the mechanical shock and vibration of the robot.} External vibration around the resonant frequency will excite large MEMS mirror vibration or even damage the mirror. To avoid the resonance effect, the MEMS mirror should avoid being actuated around the resonant frequency ($f_1, f_2, f_3$). 
\subsection{Robustness to Mirror Control Time Delay}
In ~\ref{subsec:motor_experiments}, we quantify the physical system's step response delay time at 5ms see Fig.~\ref{fig:memsstep}. In Fig.~\ref{fig:comp_vs_speed} we further quantify the effects of rotation disturbance of robot, versus compensation error, with various timing profile.

We investigate further the effects of increasing actuation time delay, of either the mirror or the mounting joint, in LiDAR SLAM simulation. We quantify the effects with SLAM odometry error, see Fig.~\ref{fig:sim_quant_blocks_delay}, Fig.~\ref{fig:sim_quant_mountains_delay} in Sec.~\ref{sec:benefit of motion compensation in SLAM}.

Further more, in Sec.~\ref{subsubsec:noisy_delay}, we further investigate the effects of time delay in our motion-compensated LiDAR Inertial Odometry pipeline, under noisy conditions. To improve on realism, we add control noise, range measurement noise, and IMU noise into our simulation.

\subsection{Robustness to Mirror Control Noise, LiDAR ToF Distance measurement Noise, and IMU Noise}
In ~\ref{subsec:motor_experiments}, Fig.~\ref{fig:comp_vs_speed} we see variations of mirror control noise under disturbances. We use a Gaussian noise model to approximate the control noise, and investigate further on increasing mirror control noise in LiDAR SLAM simulation. We quantify the effects with SLAM odometry error. See Sec.~\ref{sec:rotation_compensated_lidar_slam},Fig.~\ref{fig:various_odometry_errpr}-a).

Further, real IMU has noise, we similarly quantify the effects of increasing IMU noise in ~\ref{subsubsec:imu_noise} and Fig.~\ref{fig:various_odometry_errpr}-b). 

Furthermore, real LiDAR has ToF distance measurement noise, we similarly quantify the effects of ToF distance measurement noise in ~\ref{subsubsec:TOF noise} and Fig.~\ref{fig:various_odometry_errpr}-c).


\par \noindent To conclude, we summarize this section and everything that was described. 
\begin{itemize} 
\item We discuss the key components and characteristics of the proposed LiDAR system design, particularly focusing on the MEMS mirror. 
\item We outline a rotation compensation control algorithm that uses the MEMS mirror scanning LiDAR and sensing for compensation. We establish the coordinate systems, conversions from spherical to Cartesian coordinates, and details of the compensation process for four applications, including: general rotation compensation, 2-axes only compensation, rotational FoV stabilization and target aiming. 
\item We analyze the motion compensation control of the proposed LiDAR through real world handheld and motorized input experiments. We present the experiment data related to step response and control error versus input disturbance timing.  
\item We analyze the robustness of proposed motion compensation control in a tightly coupled LiDAR SLAM simulation. The result is presented in Sec.~\ref{sec:rotation_compensated_lidar_slam}
\item We analyze the survivability of the proposed LiDAR under robot's motion shock.

\end{itemize}
\color{black}
\section{UAV Experiment}
\label{sec:robotexp}
\begin{figure*}[!ht]
    \centering
     \begin{subfigure}{0.9\textwidth}
      \centering
      \includegraphics[width=\textwidth]{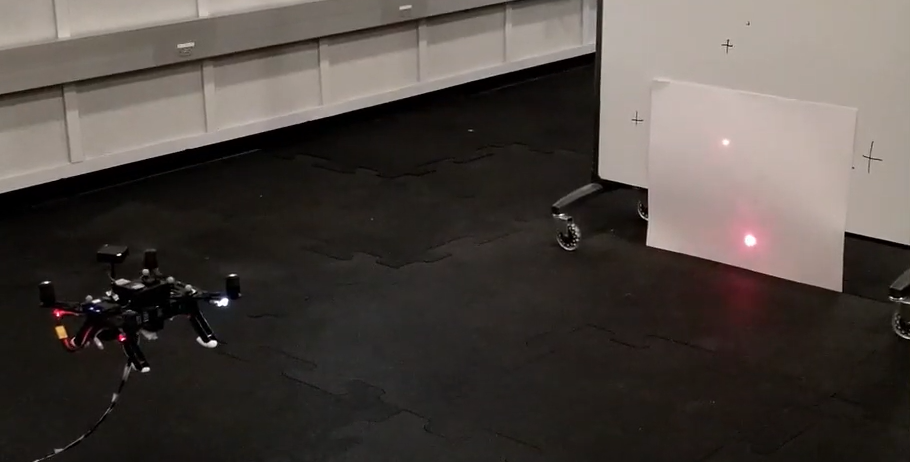}
      \caption{Our UAV setup with Intel Aero UAV and our LiDAR mounted on it. }
      \label{fig:fvsc_drone}
    \end{subfigure}%
    
    \begin{subfigure}{0.8\textwidth}
      \centering
      \includegraphics[width=\textwidth]{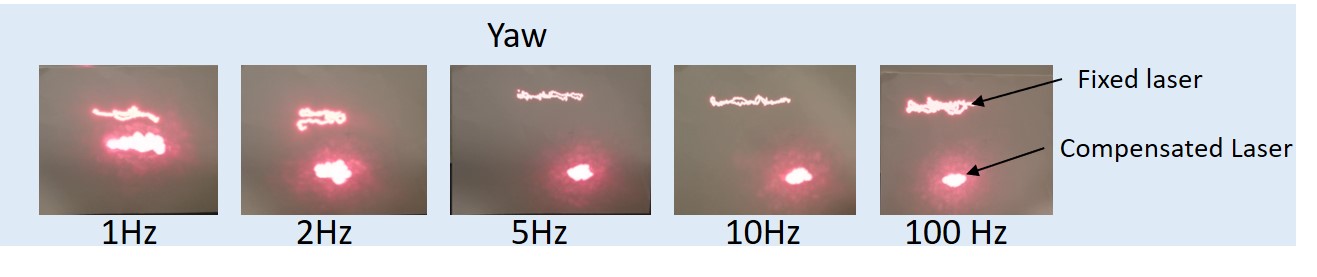}
      \caption{Effect of compensation rate on yaw rotation}
      \label{fig:fvsc1}
    \end{subfigure}%
    
    \begin{subfigure}{0.9\textwidth}
      \centering
      \includegraphics[width=\linewidth]{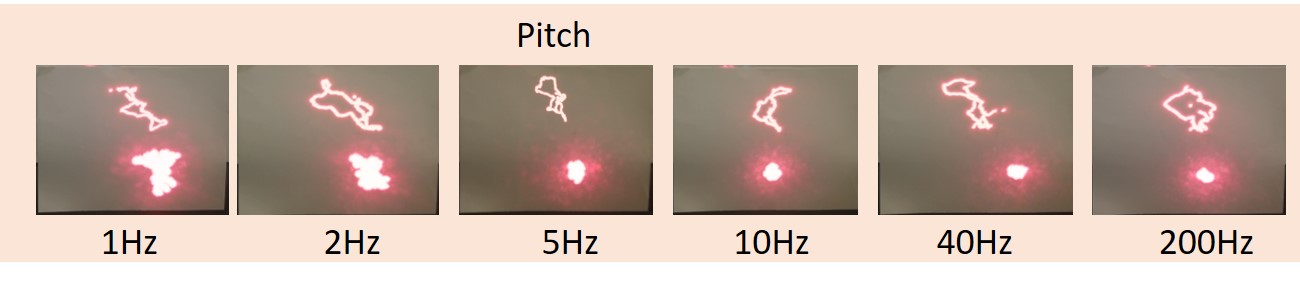}
      \caption{Effect of compensation rate on pitch rotation}
      \label{fig:fvsc2}
    \end{subfigure}%
    
    \caption{A comparison of the compensation strength versus the robot pose sampling frequencies. All the images are accumulation of 12s of UAV hovering videos. The compensation target scanning direction is a fixed direction. 
    }
    \label{fig:fvsc}
\end{figure*}

\begin{figure*} [!ht]
    \centering
    \begin{subfigure}[b]{0.9\linewidth}
      \centering
      \includegraphics[width=\textwidth]{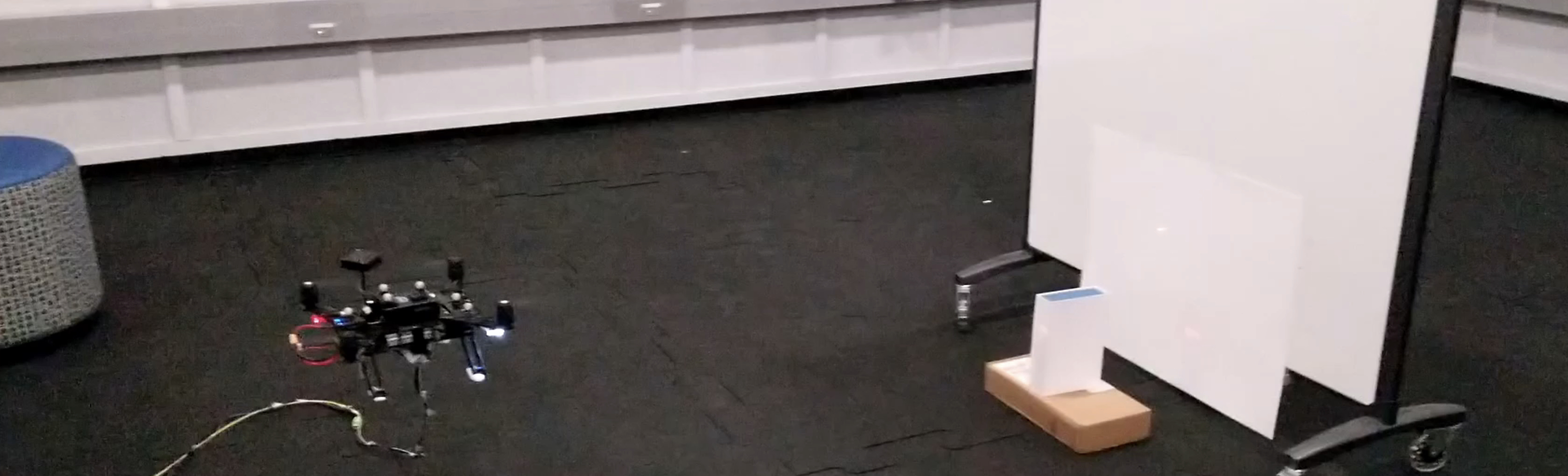}
      \caption{Our set up with UAV, LiDAR and the feature target.}
      \label{fig:uav_pointcloud}
    \end{subfigure}
    
    \begin{subfigure}[b]{0.2\linewidth}
      \centering
      \includegraphics[width=\textwidth]{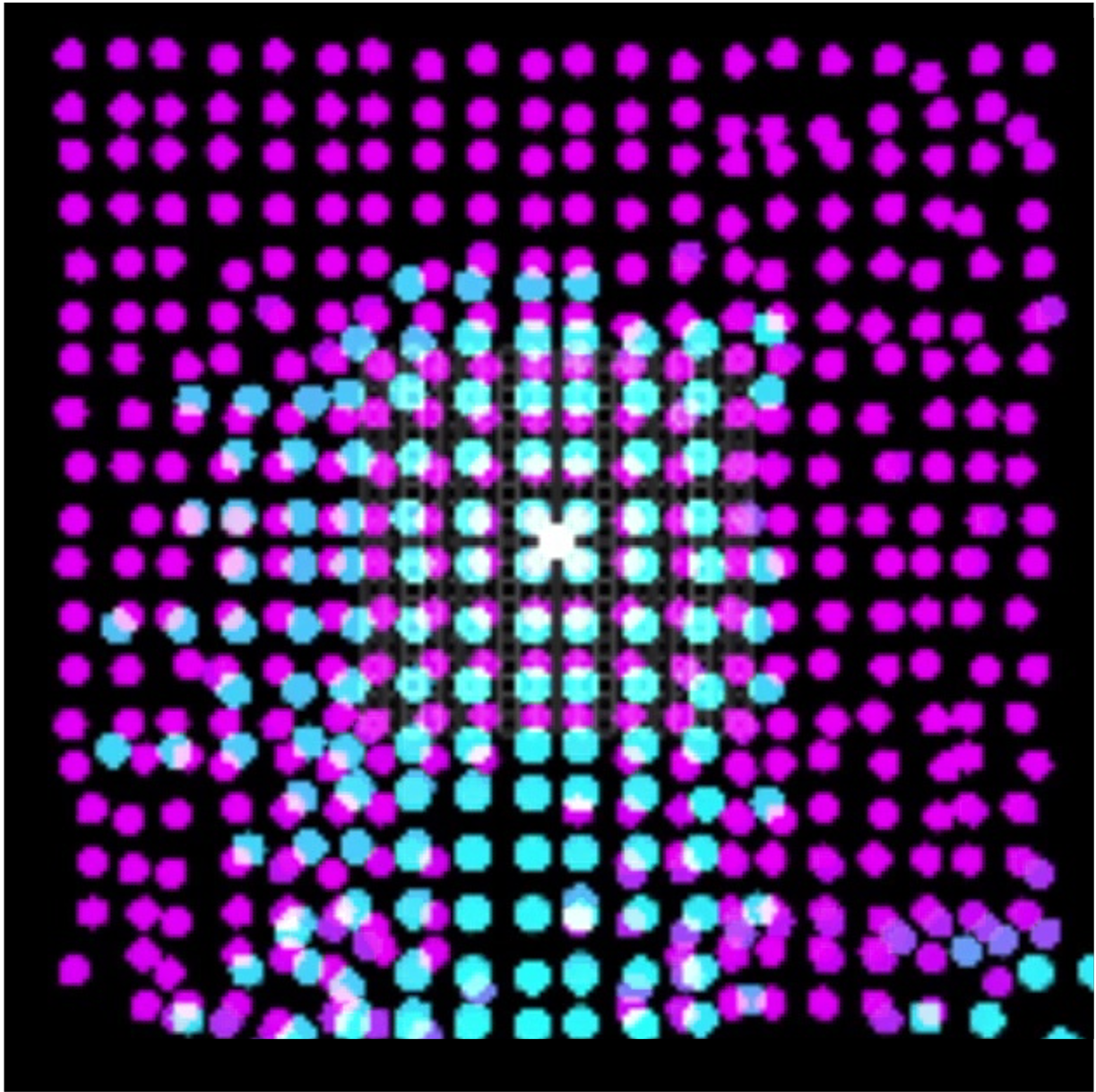}
      \caption{Uncompensated}
      \label{fig:fvpd_uc}
    \end{subfigure}%
    \begin{subfigure}[b]{0.2\linewidth}
      \centering
      \includegraphics[width=\textwidth]{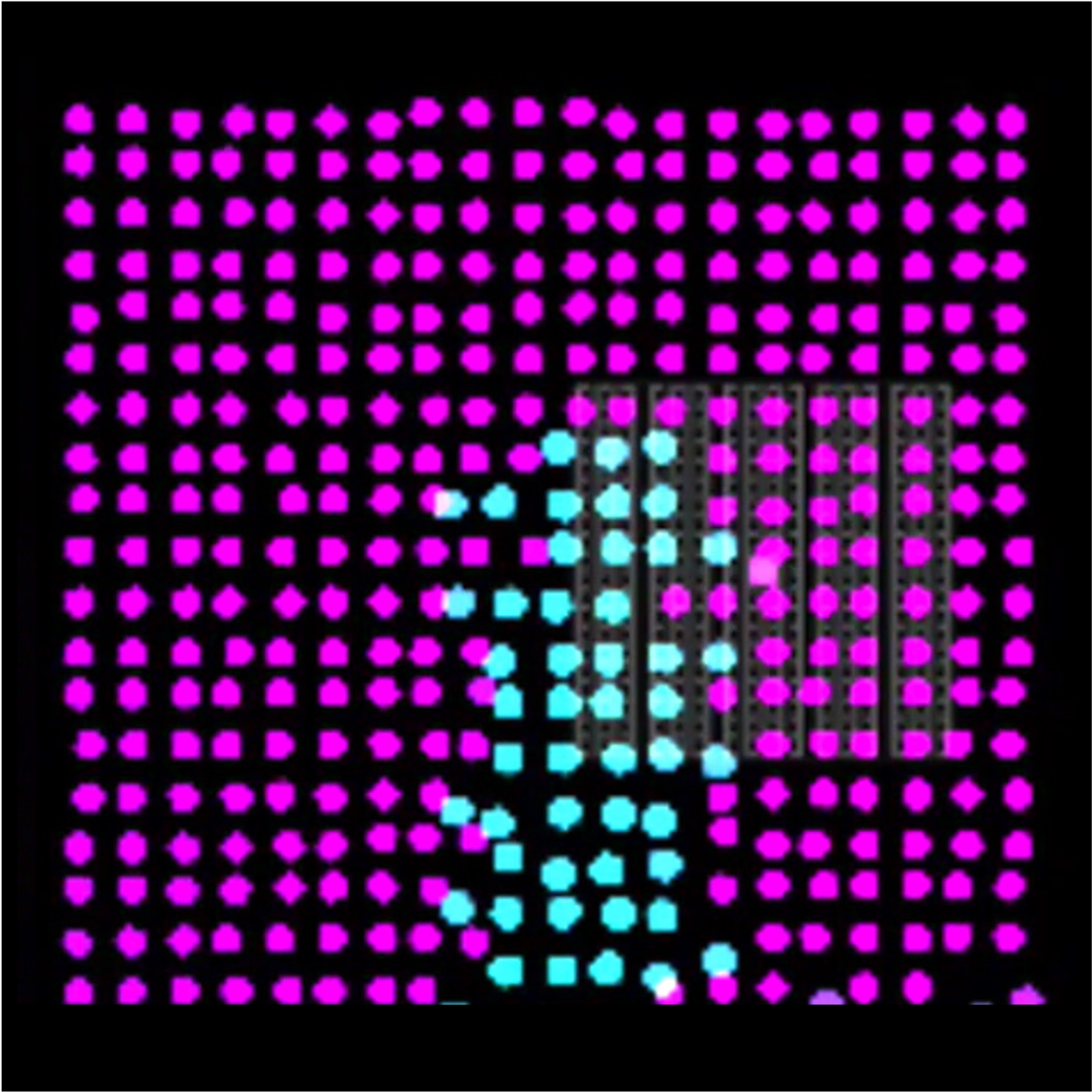}
      \caption{One frame}
      \label{fig:fvpd_1f}
    \end{subfigure}%
    \begin{subfigure}[b]{0.2\linewidth}
      \centering
      \includegraphics[width=\textwidth]{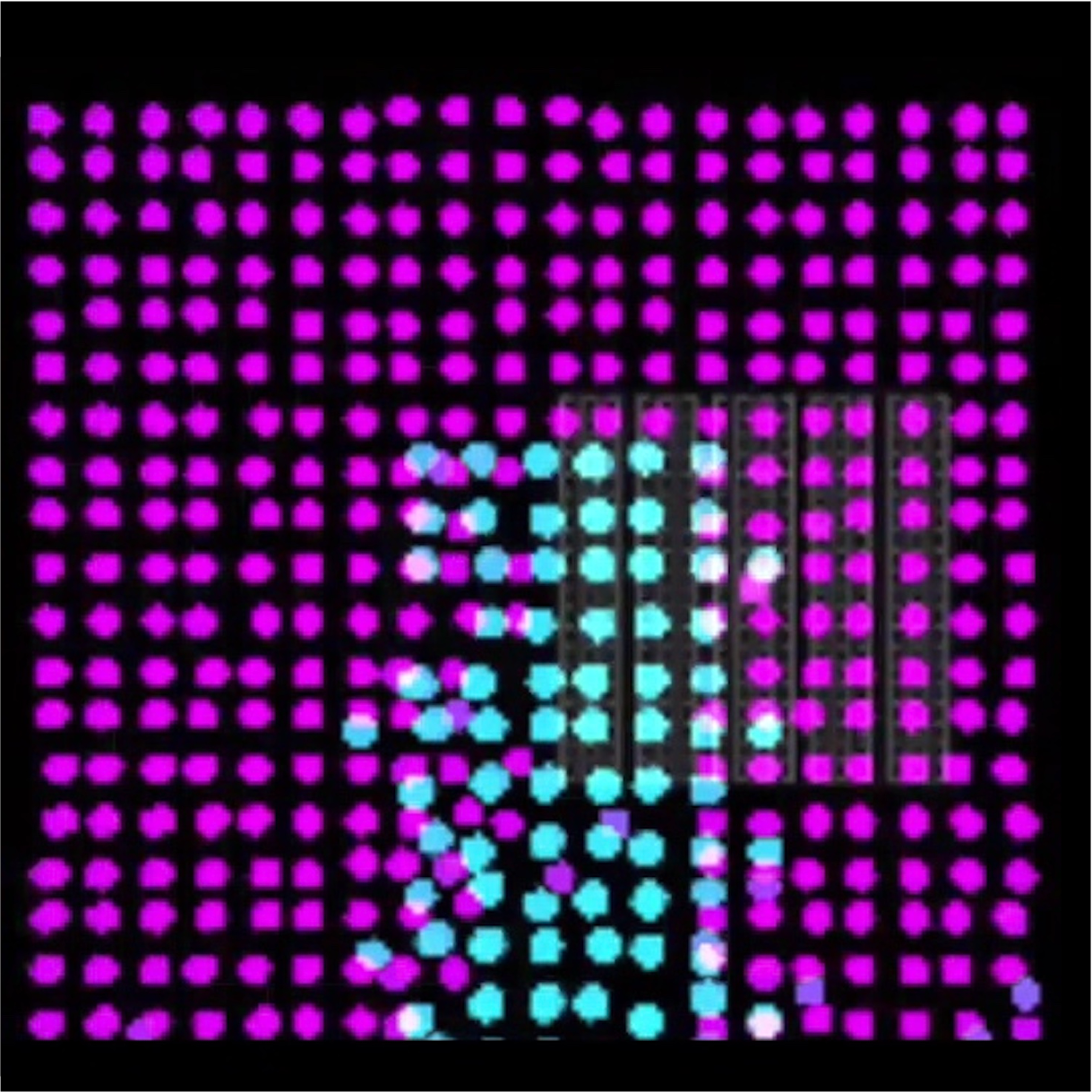} 
      \caption{50Hz sampling rate}
      \label{fig:fvpd_50hz}
    \end{subfigure}%
    \begin{subfigure}[b]{0.2\linewidth}
      \centering
      \includegraphics[width=\textwidth]{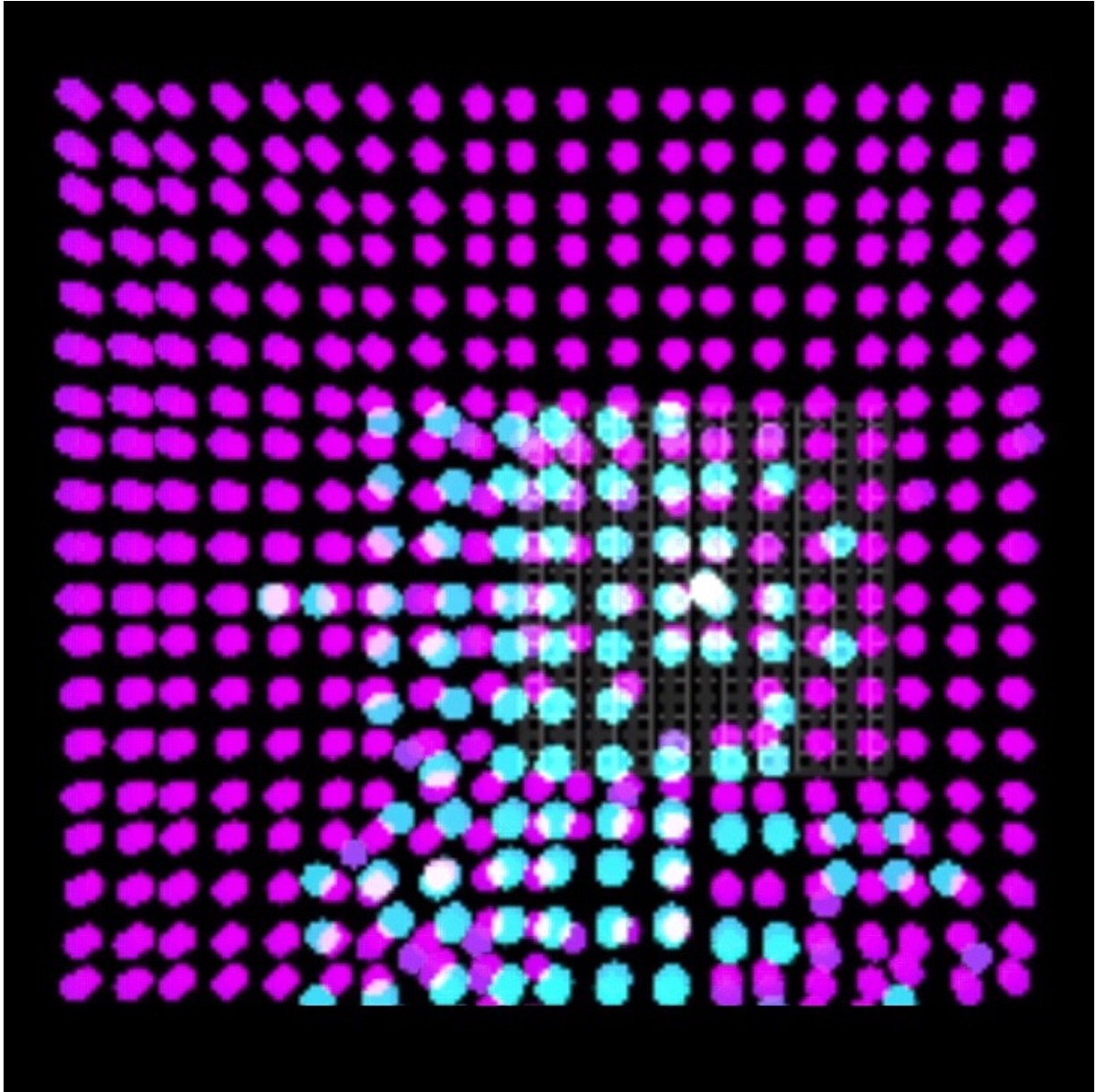}
      \caption{2Hz sampling rate}
      \label{fig:fvpd_2hz}
    \end{subfigure}%
    \begin{subfigure}[b]{0.2\linewidth}
      \centering
      \includegraphics[width=\textwidth]{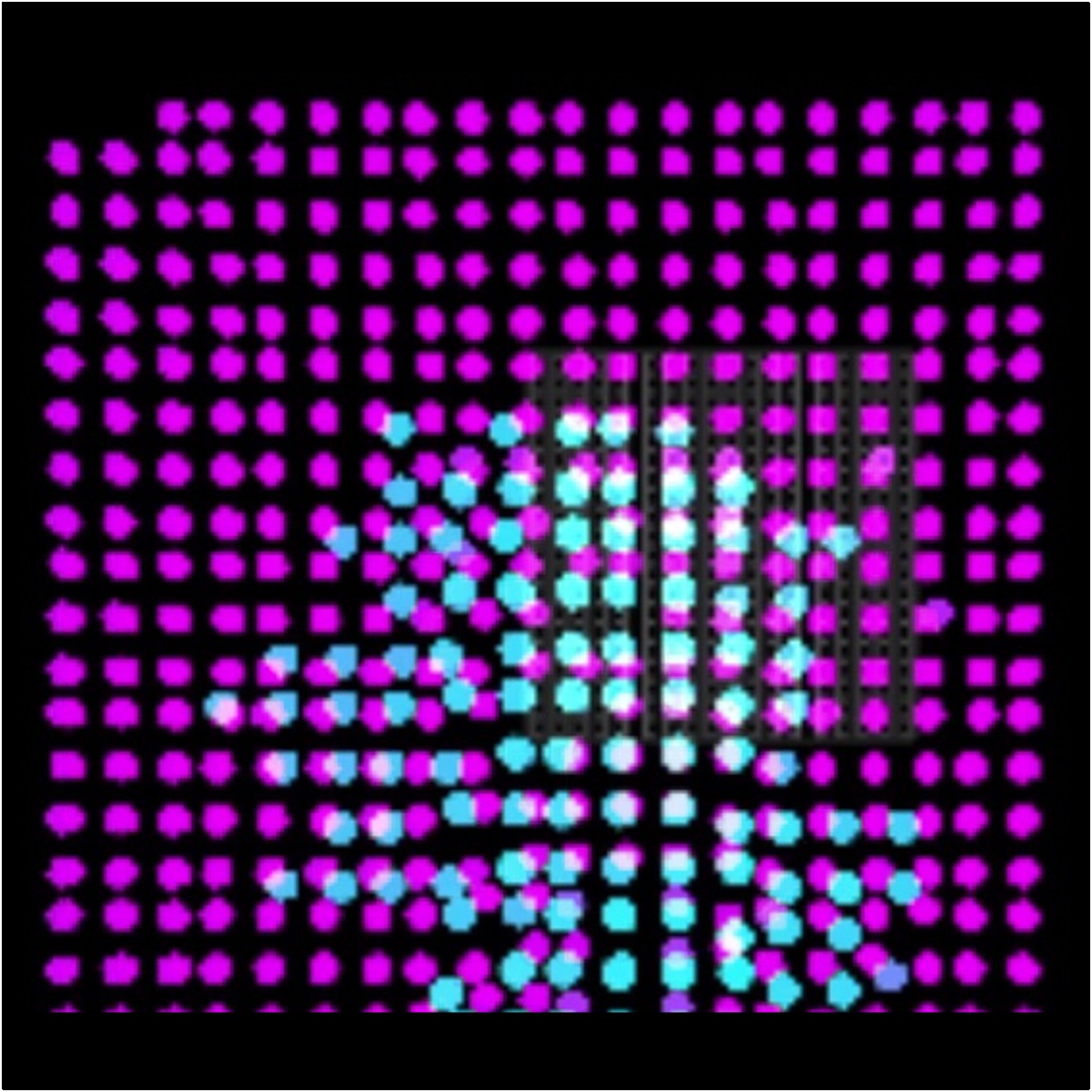}
      \caption{1Hz sampling rate}
      \label{fig:fvpd_1hz}
    \end{subfigure}%
    \caption{A comparison of the compensation strength verse the IMU sampling frequencies. The images are accumulation of 20s of point cloud video during the UAV hovering. We use a cuboidal object (as seen in Fig.~\ref{fig:uav_pointcloud}) as object of interest. 
    The width of the target increases due to compensation inaccuracy as we reduce compensation rate from 50 Hz to 1 Hz demonstrating the utility of high rates of compensation even in such static scenarios}
    \label{fig:fvpd}
\end{figure*}
Next, we demonstrated the motion compensated LiDAR by flying it on a UAV. The robot pose is from an external motion capture system that tracks the UAV. We vary the robot pose sampling rate and study its effect on the effect of compensation. The UAV is controlled to hover at a designated position with yaw/pitch rotation as motion jitter. Motion compensated LiDAR is set to compensate all the rotational motion, including the controlled rotation and the random motion disturbance. The compensated MEMS scanning laser uses a visible light, and the other visible laser is fixed at a relative higher position on the UAV, as shown in the images in Fig.~\ref{fig:fvsc1}. The target scanning direction is a fixed point on the target.

\color{black} Here, the entire scanning grid $\{\alpha_i,\beta_i,1\}$ consist of 20x20 grid pattern points. We use the aiming compensation outline in ~\ref{subsubsec:aiming}
\color{black}

We trim about 12 s videos in each experiments while the UAV is flying, and then each frames of the videos are accumulated into an image to track the motion of the UAV and the errors of the compensated scanning. 

The robot pose sampling rate is set from 1 Hz to 200 Hz to investigate its effect on the compensation results. The controlled UAV rotations are in the yaw and pitch direction. However, the actual motions cause some random motions during the flying. Point clouds are also collected when the UAV is hovering and we overlap several frames. As the robot pose sampling frequency increases from 1Hz, 2Hz to 50Hz, the width of the overlapping area shrink from 10 to 11 points at 1Hz (Fig. \ref{fig:fvpd_1hz}), to 6 points at 50 Hz (\ref{fig:fvpd_50hz}). As Fig. 14 demonstrated, the compensation frequency has clear impacts on the quality of the captured point-cloud. 

\section{Rotation Compensated LiDAR-Inertial SLAM Design}
\label{sec:rotation_compensated_lidar_slam}
\color{black} SLAM is a body of fundamental applications for visual sensors. All existing SLAM literatures reason about its odometry in the sensor's local frame, sometimes call camera frame. In this work this frame is the robot frame, with world frame orientation $\boldsymbol{R}_{robot}^w$, refer to ~\ref{subsubsec:coordinate-system}.

The basic assumption of the existing SLAM is that visual sensor readings use the robot frame with world rotation $\boldsymbol{R}_{robot}^w$ as their reference. This assumption is untrue for our sensor, because that our sensor readings use the frame with world orientation $\boldsymbol{R}_{control}\boldsymbol{R}_{robot}^w$ as their reference.

Through Sec.~\ref{subsubsec:general_rotation_compensation} to Sec.~\ref{subsubsec:aiming}, the additional none-zero rotation $\boldsymbol{R}_{control}$ orients the original scanning grid towards different directions. The existence of $\boldsymbol{R}_{control}$ breaks the basic assumption of existing SLAM.

 $\boldsymbol{R}_{control}$ must be compensated for, in order for the existing SLAM pipelines to work with our sensor. This can be done post-capture, we can use either ~\ref{subsubsec:general_rotation_compensation} or ~\ref{subsubsec:special_case} to compensate. We details the compensation later in ~\ref{subsec:the_rotation_stage}.

\color{black}

Most LiDAR odometry pipelines utilize Iterative Closest Point (ICP) to match consecutive scans and determine the rotation and translation between the poses. Any rotation of the LiDAR relative to the vehicle would cause errors in the ICP's prior. This would directly impact the quality of ICP's point-cloud registration. Although ICP can tolerate certain levels of error in its prior, in Sec.~\ref{sec:tolerable_input_level} we will show that it is far from enough when the magnitude of $\boldsymbol{R}_{control}$ input increases. 

\subsection{Motion Compensation for LiDAR SLAM}
In this simulation, we simulate a 360 degree velodyne LiDAR, that can rotate relative to the vehicle it is mounted on, by a universal joint. A universal joint has rotational DOF similar to a MEMS mirror, both limited to 2 DOF. This setup fits into the compensation framework introduced in the special case ~\ref{subsubsec:special_case}.  In this section, we will demonstrate in simulation that such rotation introduces error in an off-the-shelf LiDAR SLAM pipeline. Additionally, We propose a general method to incorporate such rotation into consideration when performing LiDAR-related SLAM. We demonstrate the effectiveness of the framework in a Rotation Compensated LiDAR-Inertial Odometry and Mapping package, which is publicly available on  ~\href{https://github.com/yuyangch/LIO-SAM_rotation}{Github}.

\begin{figure}
\centering
 \includegraphics[trim={0 0cm 0 0cm},clip,width=.95\columnwidth]{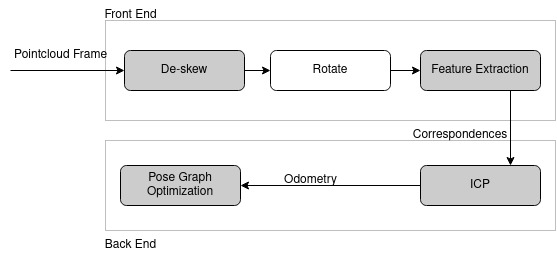}
 \caption{Illustration of our rotating LiDAR SLAM augmentation (existing modules are shown in grey)}
 \vspace{+0pt}
 \label{fig:rotating_lidar_slam_pipeline}
\end{figure}
For the ease of integration, our framework proposal does not make large edits in the existing paradigm. It only adds a ``rotate" stage right after the de-skew stage in the front end, and before feature extraction stage. This edition can be easily integrated with existing pipeline and future designs. The rotate stage does one single operation, it rotates the de-skewed point cloud according to the control rotation input to the LiDAR. Our workflow block diagram in shown in Fig.~\ref{fig:rotating_lidar_slam_pipeline}.
\subsection{The Rotation Stage}
\label{subsec:the_rotation_stage}
    The purpose of this stage of the pipeline, is to rotate the captured LiDAR frame, to a correct position, relative to the LiDAR's base frame of reference. (In this work, the LiDAR's base frame is identical to the vehicle's body frame.)

    Let the Lidar's base frame have world  rotation \color{black}$\boldsymbol{R}_{robot}^w \in SO(3)$\color{black}. 

    In a traditional LiDAR that doesn't rotate, all points received in a LiDAR frame are position relative to the LiDAR's base frame, with world rotation \color{black}$\boldsymbol{R}_{robot}^w$\color{black}. However, this assumption is incorrect for our device, where the LiDAR frame is positioned relative to the frame with rotation $\boldsymbol{R}_{control}\boldsymbol{R}_{robot}^w$.

    The LiDAR's head can rotate \color{black}$\boldsymbol{R}_{control} \in SO(3)$\color{black}, relative to its base. This rotation is restricted to azimuth $\beta$ and elevation directions $\alpha$. \color{black} note that in here we analyze a more generalize, special case compensation~\ref{subsubsec:special_case}, but it can be easily extend to full $SO(3)$ compensation~\ref{subsubsec:general_rotation_compensation}\color{black},

    When a LiDAR frame is received, we take the most recently known rotation $\alpha,\beta$, in this case the most recent known command rotation, and converts them into a rotation matrix:
    \begin{gather}
    \label{eq:lidar_head_rotation_matrix}
\color{black}\boldsymbol{R}_{control}\color{black}
 =
   \begin{bmatrix}
   \cos{\beta} & -\sin{\beta}&0 \\
    \sin{\beta} & \cos{\beta} & 0 &\\
   0 & 0 & 1
   \end{bmatrix}
  \begin{bmatrix}
   \cos{\alpha} & 0 & \sin{\alpha} \\
    0 & 1 & 0 &\\
   -\sin{\alpha} & 0 & \cos{\alpha} 
   \end{bmatrix}
\end{gather}

    And applies the rotation to each point $p \in \mathbb{R}^3 $ in the frame point-cloud:

    \begin{gather}
    \label{eq:rotated_pointcloud}
p_{rotated}
 = \color{black}\boldsymbol{R}_{control}\color{black}p
\end{gather}
    The rotated point-cloud $p_{rotated}$ now locates at the correct position, relative to the LiDAR's base frame, with world rotation $\boldsymbol{R}_{robot}^w$. The basic assumption of traditional SLAM are now met.

\subsection{Evaluation}
\label{sec:SLAM-eval}
Now we evaluate the sensor in simulation to answer a few questions. First, we want to compare traditional LiDAR SLAM and our Motion Compensated SLAM in terms of the handling change in mirror/universal joint orientation magnitude. Next we investigate the effect of noise in the mirror's orientation (say through a faulty IMU or other sensor) on the robustness of our pipeline. We also show the degree to which our pipeline can tolerate such noise. 

The proposed SLAM framework should be expected to function, even when the LiDAR users employ control policies that rotate its FoV significantly frame-to-frame. This is unlike the scenario of running a active stabilization control policy proposed in ~\ref{sec:benefit of motion compensation in SLAM}, where frame-to-frame variation is minimal. Therefore, in this evaluating section, we use control policy that samples random LiDAR rotation control input from Gaussian distributions at high frequency. 

We choose LIO-SAM as the traditional SLAM package to compare against, and built our Motion Compensation framework into it, and open-source it on github. LIO-SAM has all the signature point-cloud processing stages shown in Fig.~\ref{fig:rotating_lidar_slam_pipeline}. It is relatively new and has good SLAM accuracy performance versus State-of-the-Art. We hope through the open-source code we can demonstrate to the community an example of incorporating our framework.

For Odometry error evaluation, We calculate Average Translation Error (ATE) which is defined by the KITTI benchmark ~\cite{geiger2012we}:
\begin{gather}
    \label{eq:kitti_benchmark_eqn}
E_{trans}(\mathcal{F})
 = \frac{1}{|\mathcal{F}|} \sum_{i,j \in \mathcal{F}}\|\hat{T}_j\hat{T}_i^{-1}(T_{j}T_i^{-1})^{-1}\|_{2}
\end{gather}
Where $\mathcal{F}$ is a set of frames $(i,j)$, $T$ and $\hat{T}$ are the estimated and true LiDAR poses respectively.


\subsubsection{Experiment Set Up}
\begin{figure*}
\centering
 \includegraphics[width=.9\textwidth]{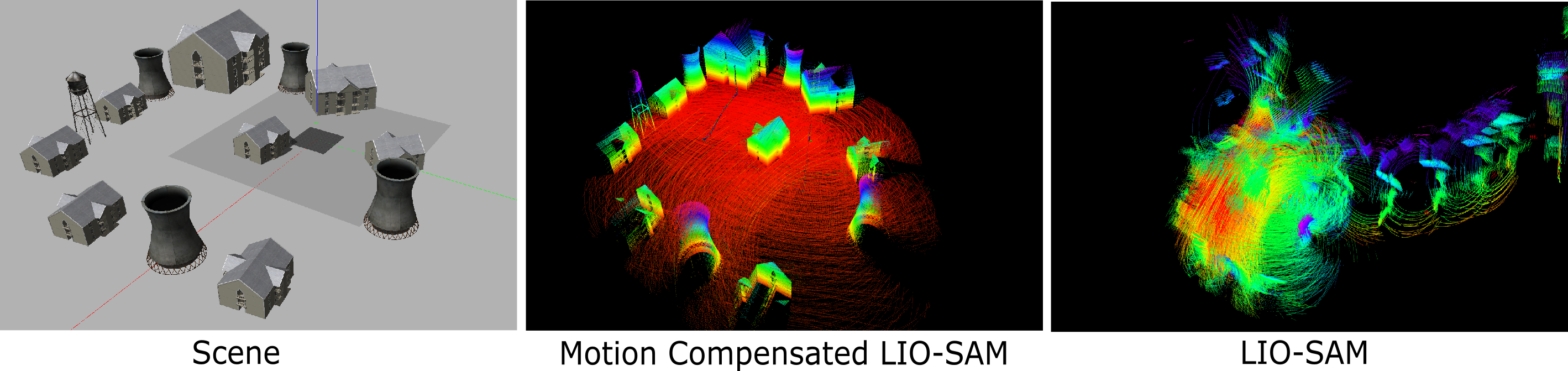}
 \caption{Illustration of a) our simulator environment, b) mapping results on LIO-SAM with our motion compensation c) mapping results on stock LIO-SAM. It is worth noting that the pointcloud generated from the simulation has rotation with respect to the frame with world rotation $\boldsymbol{R}_{control}\boldsymbol{R}_{robot}^w$, which breaks any traditional visual SLAM's assumption, see Sec.~\ref{sec:rotation_compensated_lidar_slam}. It has significant frame-to-frame FoV variations (See Sec.~\ref{sec:SLAM-eval}), which is difficult for any un-compensated, traditional SLAM to handle.} 
 \vspace{0pt}
 \label{fig:px4_gazebo_simulator}
\end{figure*}

A simulation study is setup in robotics simulator Gazebo, where a LiDAR with similar sensor characteristics to a Velodyne VLP-32 is mounted on a simulated drone. Further, the LiDAR can rotate in azimuth and elevation via a universal joint. The simulated drone iris, is from the PX4's simulation package. Its onboard IMU have noise added to it according to a noise model outlined in Kalibr ~\cite{furgale2013unified}. The point-cloud messages from the LiDAR, as well as the IMU messages from the drone, are passed into robotics middleware ROS, where the proposed LiDAR SLAM package runs.
The drone is commanded to flight in a diamond waypoint pattern, around a enviornment with different types of resident buildings.

The proposed LiDAR-Inertial SLAM package builds on top of LIO-SAM, which employs the powerful PGO backend GTSAM ~\cite{dellaert2012factor}. We incoporate the compensation described in \ref{subsec:the_rotation_stage} into LIO-SAM, which we refer here on in as Motion Compensated
LIO-SAM. Naturally, we will compare SLAM performance of Motion Compensated LIO-SAM, against the stock version of LIO-SAM. See Fig.~\ref{fig:px4_gazebo_simulator}. To control the orientation of the universal joint, angular commands in $\alpha,\beta$, in degrees, are input to the mirror.

\subsubsection{Level of mirror control orientation magnitude tolerable by an unmodified pipeline vs our system}
\label{sec:tolerable_input_level}
\begin{figure*}
\centering
 \includegraphics[width=.9\textwidth]{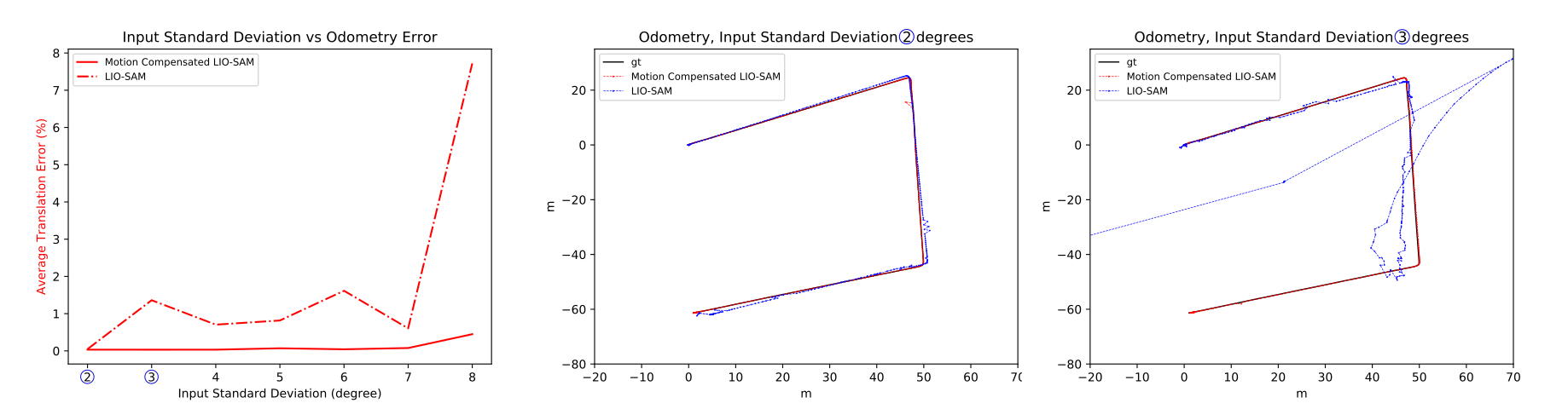}
 \caption{ \color{black}  a) Mirror control magnitude and odometry error, Our Motion Compensated LIO-SAM vs LIO-SAM . As mirror control magnitudes increase, the unmodified LIO-SAM fails completely. b) At 2 degrees  standard deviation, Our Motion Compensated LIO-SAM outperforms LIO-SAM c) At 3 degrees standard deviation, degree threshold and beyond, Our Motion Compensatedd LIO-SAM perform normally while the stock LIO-SAM completely fails}

 \label{fig:mirror_control_vs_odometry_error}
\end{figure*}

The two angular commands are sampled from 1-d Gaussian distributions with standard deviation of various degrees, at 10 Hz. Odometry error vs command rotation's gaussian standard deviation is plotted in Fig.~\ref{fig:mirror_control_vs_odometry_error}. 
An Gaussian distribution with 8 degree standard deviation generate input angle within +-,8,16,24 degrees, 68,95 and 99.7 percent of the time respectively. Therefore 99.7 percent of the time, angular input span a range of 48 degrees. 

In short, by considering mirror rotation, the system can tolerates angular input that span 48 degrees. In contrast, without mirror rotation information, the system can only tolerate angular input that span 12 degrees.

Even in the cases where the input spans less than 12 degrees, by considering mirror rotation, SLAM quality improves in comparison.

\begin{figure*}
\centering
 \includegraphics[width=.9\textwidth]{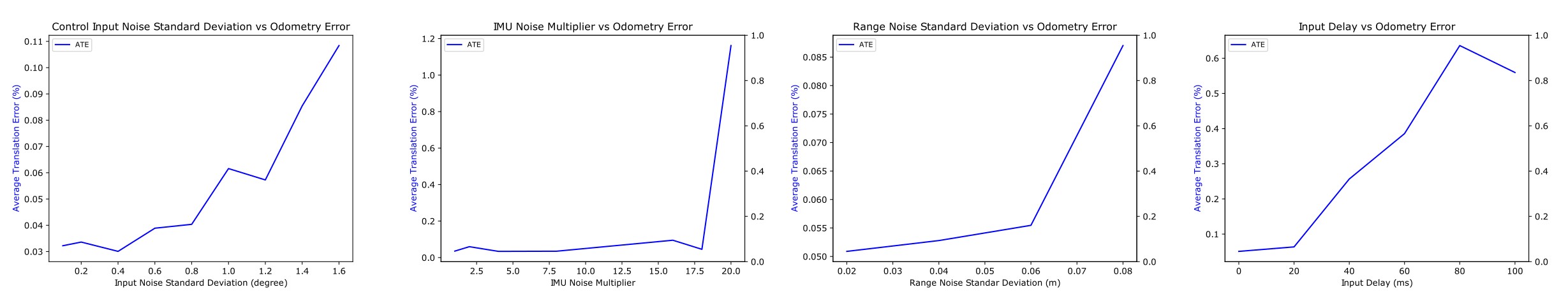}
 \caption{ \color{black}  a) Mirror control noise and odometry error. As the mirror control noise increases, the odometry error also increases. Our pipeline fails after noise standard deviation exceeds 1.6 degree. b) Imu noise and odometry error. As IMU noise factor increase, odometry error also increases. SLAM failure in our pipeline occurs at factor 20. c) As range measurement noise increase, odometry error also increases d)As input delay increases, odometry error also increase. Our system can tolerate input delay of 30ms. }

 \label{fig:various_odometry_errpr}
\end{figure*}

\subsubsection{Level of mirror control noise tolerable   }

The two angular commands are sampled from 1-d Gaussian distributions with standard deviation of 3 degrees, at 10 Hz. We use our proposed Motion Compensated LIO-SAM here. 

Aditionally, Noise rotations in both azimuth and elevation are added on top of each channel. Odometry error vs command rotation's noise Gaussian standard deviation is plotted in Fig.~\ref{fig:various_odometry_errpr}-a).
The system can tolerate mirror input control noise up to 1.6 degree standard deviation, which spans 9.6 degree.

As Fig.~\ref{fig:comp_vs_speed} shows, when 1.8 degrees of disturbance from the robot body is generated, there is approximately 10 percent or ~+-0.18 degrees of peak actuation error, and approximately 3 percent, or ~+-0.05 degrees of actuation error off peak. Assuming in the worst case a robot generate 18 degrees of disturbances from the robot body, this translate to 1.8 degrees of peak actuation error and 0.5 degrees of actuation error off-peak. Using a Gaussian error profile, we have standard deviation at 1.8/3=0.6 degrees, which is within the noise level that the motion-compensated LIO-SAM can tolerate.


    
\subsubsection{Level of IMU noise tolerable}
\label{subsubsec:imu_noise}
The two angular commands are sampled from 1-d Gaussian distributions with standard deviation of 3 degrees, at 10 Hz.We use our proposed Motion Compensated LIO-SAM here.
Additionally, IMU noise are added on top of IMU output, according to an IMU noise model outlined in kalibr ~\cite{furgale2013unified}. The model is based on ~\cite{ma1998ieee}\cite{trawny2005indirect}\cite{crassidis2006sigma}. The base set of noise parameters are outline in the following table.  

\setlength{\tabcolsep}{2pt}
\begin{tabular}{ | c| c| c | }
  
  \hline
  Parameter & Unit & Value \\
  \hline
  Gyroscope Noise Density& $\frac{rad}{s} \frac{1}{\sqrt{Hz}} $&   3.394e-4\\ 
  \hline
  Gyroscope Random Walk& $\frac{rad}{s^2} \frac{1}{\sqrt{Hz}} $ &    3.879e-05\\ 
  \hline
  Gyroscope Turn On Bias Sigma & $\frac{rad}{s} $ & 8.727e-3 \\ 
  \hline
  Accelerometer Noise Density& $\frac{m}{s^2} \frac{1}{\sqrt{Hz}} $&     4e-3\\ 
  \hline
  Accelerometer Random Walk& $\frac{m}{s^3} \frac{1}{\sqrt{Hz}} $ &     6e-3\\ 
  \hline
  Accelerometer Turn On Bias Sigma & $\frac{m}{s^2} $ & 0.196\\ 
  \hline
 
\end{tabular}

\rule{0pt}{4ex} 

The above set of noise parameters are multiply by several factors and their relationship with odometry error are shown in Fig.~\ref{fig:various_odometry_errpr}-b). The following set of noise parameters remains constant.

\rule{0pt}{4ex}

\begin{tabular}{ | c| c| c | }
  
  \hline
  Parameter & Unit & Value \\
  \hline
  Gyroscope Bias Correlation Time & $s $ & 1000 \\ 
  \hline
  Accelerometer Bias Correlation Time & $s $ & 300 \\ 
  \hline
 
\end{tabular}

\rule{0pt}{4ex} 

\emph{In conclusion, the modified pipeline can tolerate IMU noise factor of up to 18.}


\subsubsection{Level of LiDAR ToF Distance Measurement Noise Tolerable}
\label{subsubsec:TOF noise}
Velodyne's VLP-16 has a Gaussian distance measurement noise profile, with 0 mean,  .005-.008 standard deviation~\cite{glennie2016calibration}.
We simulate increasing Gaussian distance measurement noise versus Odometry error.

The two angular commands are sampled from 1-d Gaussian distributions with standard deviation of 3 degrees, at 10 Hz. We use our proposed Motion Compensated LIO-SAM here.

Additionally, we add Gaussian distance measurement noise of 0 mean and varying standard deviation, ranging from 0.02-0.08, which is about 4-10 times of distance noise from a commercially available VLP-16 LiDAR. The result can be seen in Fig.~\ref{fig:various_odometry_errpr}-c).



\subsubsection{Level of actuation delay tolerable under noisy condition}
\label{subsubsec:noisy_delay}
Our modification to LIO-SAM requires timely reporting of actuator joint/MEMS position. Actuation delay can therefore impact the odometry accurary. 

Here we evalute motion-compensated LIO-SAM's odometry accuracy with increasing actuation delay.

Different than~\ref{sec:benefit of motion compensation in SLAM}, to increase realism, we add the above mentioned mirror control noise, IMU noise, LiDAR distance measurement noise. 

The mirror input control Gaussian noise is set at
0 mean and 0.3 degree standard deviation. 

The IMU noise is set at 3X of base IMU noise level mentioned in ~\ref{subsubsec:imu_noise}.

The LiDAR distance Gaussian measurement noise is set at 0 mean and 0.008 degrees standard deviation, similar to that of VLP-16.

The two angular commands are sampled from 1-d Gaussian distributions with standard deviation of 3 degrees, at 10 Hz. We use our proposed Motion Compensated LIO-SAM here.
 The result can be seen in Fig.~\ref{fig:various_odometry_errpr}-d).
 


\color{black}
\section{Limitations and Conclusions}
We have designed an adaptive lightweight LiDAR capable of reorienting itself. We have demonstrated the benefits of such a LiDAR in simulation as well as experiment. We have demonstrated in experiment image stabilization in hardware using an onboard IMU. We have also demonstrated viewing an object of interest using this LiDAR through external robot pose feedback. Please see the supplementary material of this paper for some MEMS-related details, including analysis of robot motion shock on the MEMS as well as preliminary point cloud stitching. We also explain how such a sensor can reduce sensing uncertainty. Finally, our accompanying video shows our experiments in action. 

We would also like to acknowledge limitations of our study. 
\begin{itemize}

\item We have indirectly compared to software methods using compensation {delay}. This is because, compared to hardware-compensation, any software-compensation will add delay, and therefore delay is a fundamental metric for hardware-software comparison. For future work we will directly compare with software compensation methods. 

\item Our design requires the robot to connected to the heavier sensing components using a tether. This limits the fly range and the detection FoV of the system. Although removing the tether restriction is left to future work, we believe that our design is capable of advancing sensing in microrobots significantly, and will help our community in designing microrobots in the future. 

\item All our results (using IMU as well as Vicon motion capture) are indoor results. We hope to perform future experiments with outdoor effects such as wind.

\item In our current system design, there are implementation bottlenecks that limit compensated bandwidth. These are caused by the MEMS mirror and by the signal processing. Tightly coupled on-board designs can reduce these.
\item In our current system design, manufacturing and material constraints have limited current MEMS scanners' FoV and speed, making them more suitable for small-sized and lightweight LIDAR applications.
\color{black}
\end{itemize}

In conclusion, through simulation and a prototype implementation we realize our design shown in Fig \ref{fig:futuristic}. We have shown, in simulation and on real hardware experiments, that hardware-compensation using a MEMS mirror improves both reconstruction and mapping. In particular, microrobots which suffer from heavy vibration and motion jitter (such as flapping-wing MAVs~\cite{robobees-sciam13}) can benefit greatly from the motion compensated MEMS mirror scanning LiDAR for stabilized scene capture. Finally, over the long term, we believe that our design methodology can decouple robot and sensor geometry, greatly simplifying robot perception. 

\section{Acknowledgments}
The UF authors wish to thank and acknowledge partial support from the ONR from grants N00014-18-1-2663 and N00014-23-1-2429 as well as the NSF from grant 1942444. The UB authors wish to thank and acknowledge partial support from grants NSF-1514395 and NSF-1846320.


    
    


\printbibliography


\section*{Supplementary materials}

\subsection*{Motion Compensation Controller Design}

Those algorithms provide motion-compensated MEMS scanning in steady-state. However, as both the IMU and the MEMS mirror have limited response bandwidths, the residue and the speed of the motion compensation may get worse as the frequency of motion jitter increases. Also, the MEMS mirror has a limited scanning range, so the range of motion compensation is also limited. Since the motion compensation system is an open-loop system, a compensating stage $H_g(s)$ can be added to the microcontroller to increase the performance of the system.

\begin{figure}
    \centering
    \includegraphics[width=\linewidth]{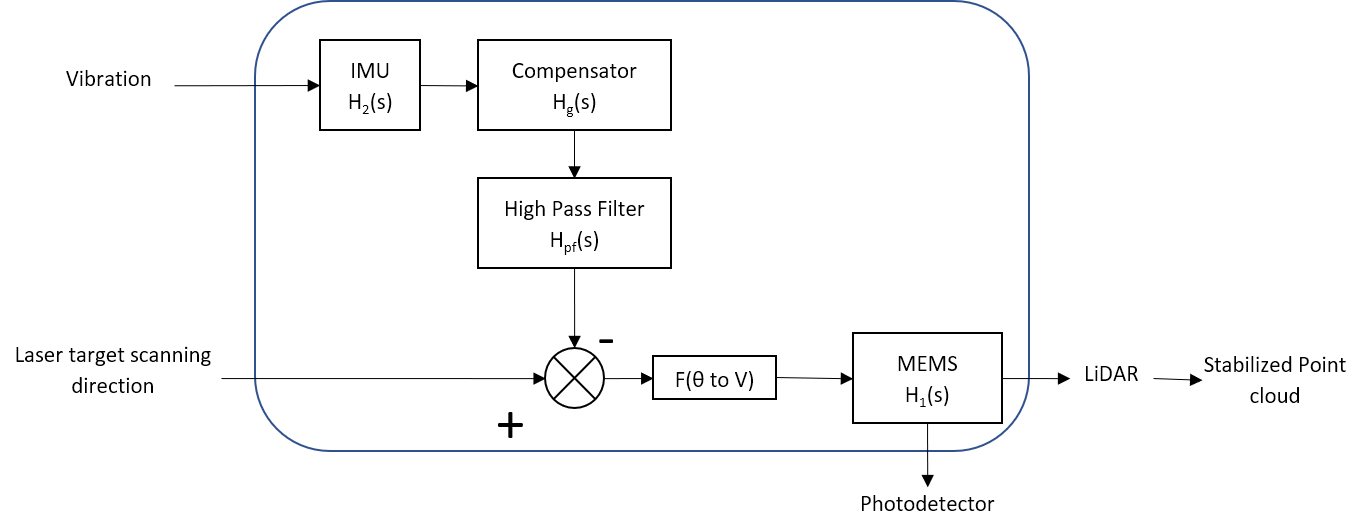}
    \caption{The block diagram of the open-loop motion compensation system.}
    \label{fig:block}
\end{figure}

A simplified block diagram of the open-loop motion compensation system is shown in  Fig.~\ref{fig:block}, where $H_1(s)$ denotes the MEMS mirror tip-tilt model given in Eq.~\ref{H1s}, $H_2(s)$ is the IMU model, $H_g(s)$ is the compensating components, $H_{pf}(s)$ is an optional high-pass filter and $F$ is the model for the MEMS mirror driver.

\begin{figure}
     \centering
    \includegraphics[width=\linewidth]{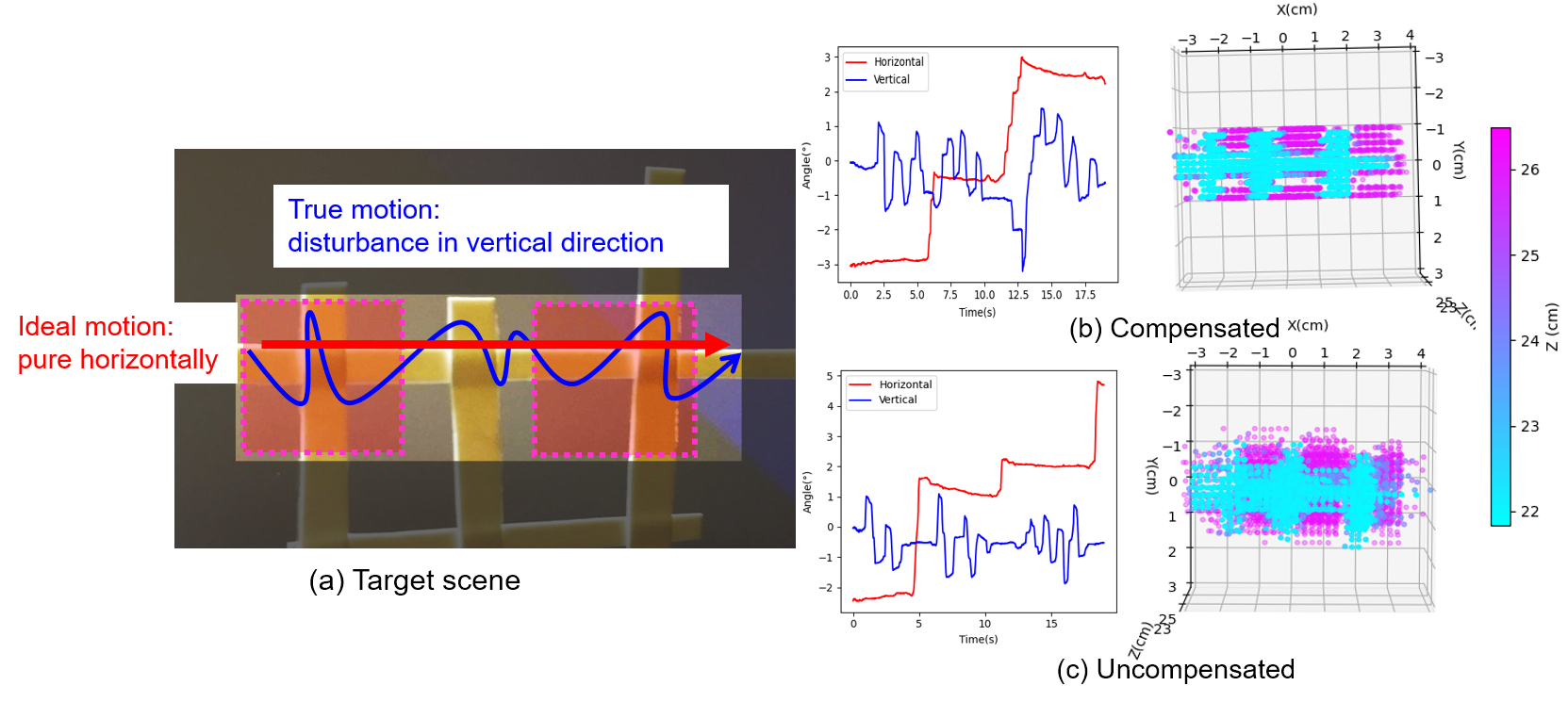}
    \caption{Compensated point cloud stitching and experiment and its results. (a) The target scene placed at 22 cm from the LiDAR. The red line is the ideal motion, the blue line qualitatively depicts the type of experiment we performed, with vertical disturbances only. The graphs show the actual, measured  disturbances. (b) With compensated scanning, the recorded motion and the generated point cloud show sharper depth edges than that with (c) with uncompensated scanning. }
    \label{fig:stitching}
    \vspace{-10pt}
\end{figure}

\subsection*{Stitching experiments with translation}

Here, we performed point-cloud stitching as that sensor moves along an object. The target scanning area is along the horizontal paper-cut figure shown in the highlighted area in the Fig.~\ref{fig:stitching} (a). The blue line is an example of the true motion with disturbance only existing in the vertical direction. As the LiDAR rotates from in the horizontal direction from left to right, it is expected to collect the best point cloud covering the highlight area of the object only. The result of compensated scanning and uncompensated scanning are shown in Fig.~\ref{fig:stitching} (b) and (c) on the right side, with their measured motions on the left side. Please find other supplementary materials in the accompanying video.

\begin{IEEEbiography}[{\includegraphics[width=1in,height=1.25in,clip,keepaspectratio]{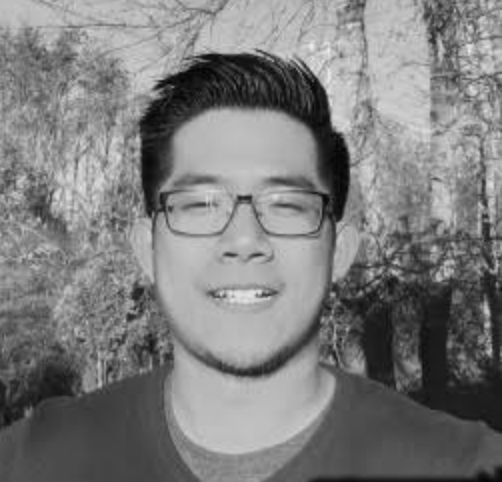}}]{Yuyang Chen}
received his Bachelors of Science in Electrical Engineering from Stony Brook University. He earned Ph.D. of Computer Science from University at Buffalo, under the supervision of Prof. Dantu.
\end{IEEEbiography}

\begin{IEEEbiography}[{\includegraphics[width=1in,height=1.25in,clip,keepaspectratio]{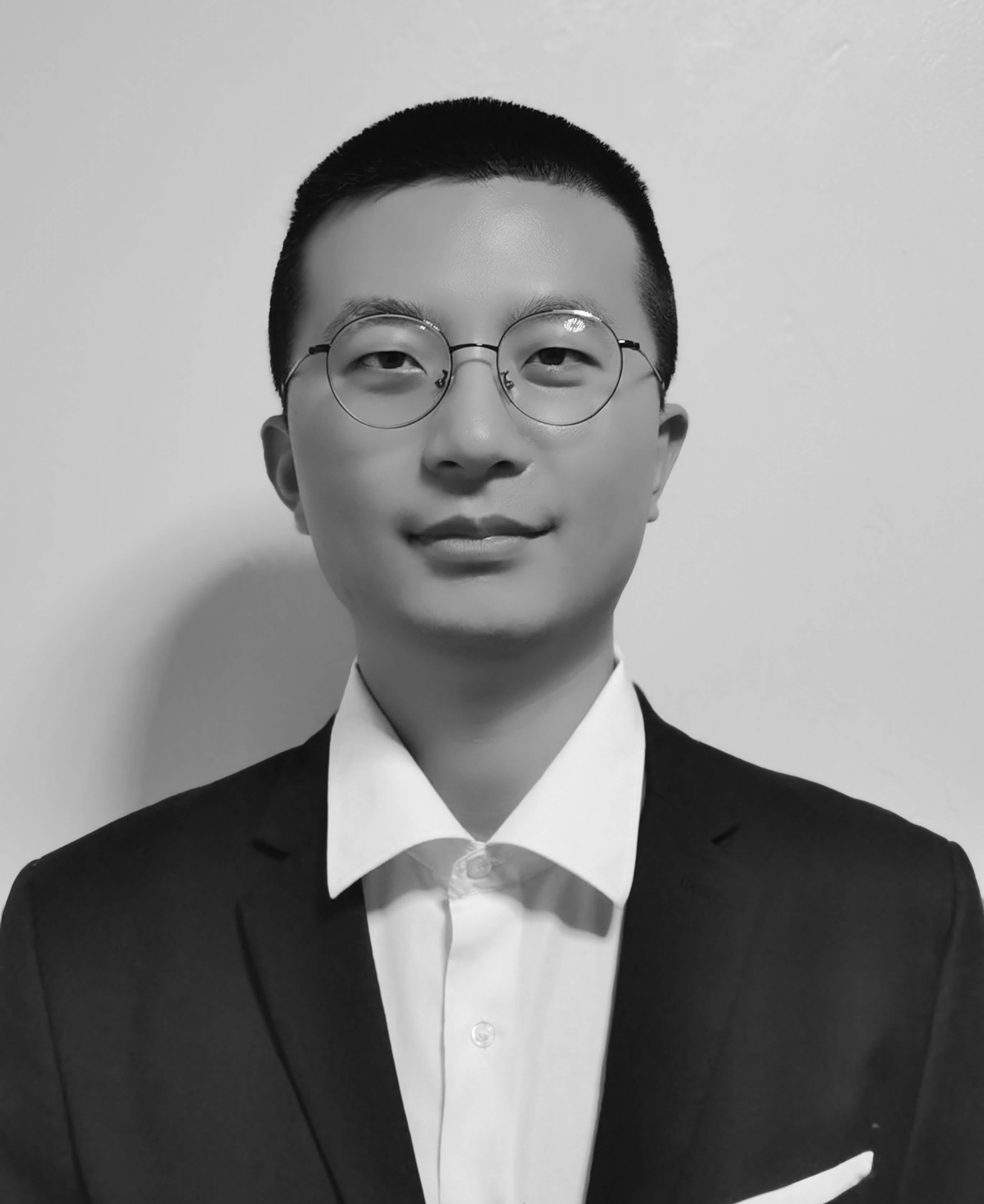}}]{Dingkang Wang}
received the B.E. degree in mechanical engineering from Jilin University, China, the Ph.D. degree in Electrical Engineering from the University of Florida, Gainesville, USA. He is now a LiDAR and computer vision engineer in the autonomous driving industry. His research interests are how to use machine learning, multi-sensor fusion, and generative AI algorithms to help autonomous driving tasks
\end{IEEEbiography}

\begin{IEEEbiography}[{\includegraphics[width=1in,height=1.25in,clip,keepaspectratio]{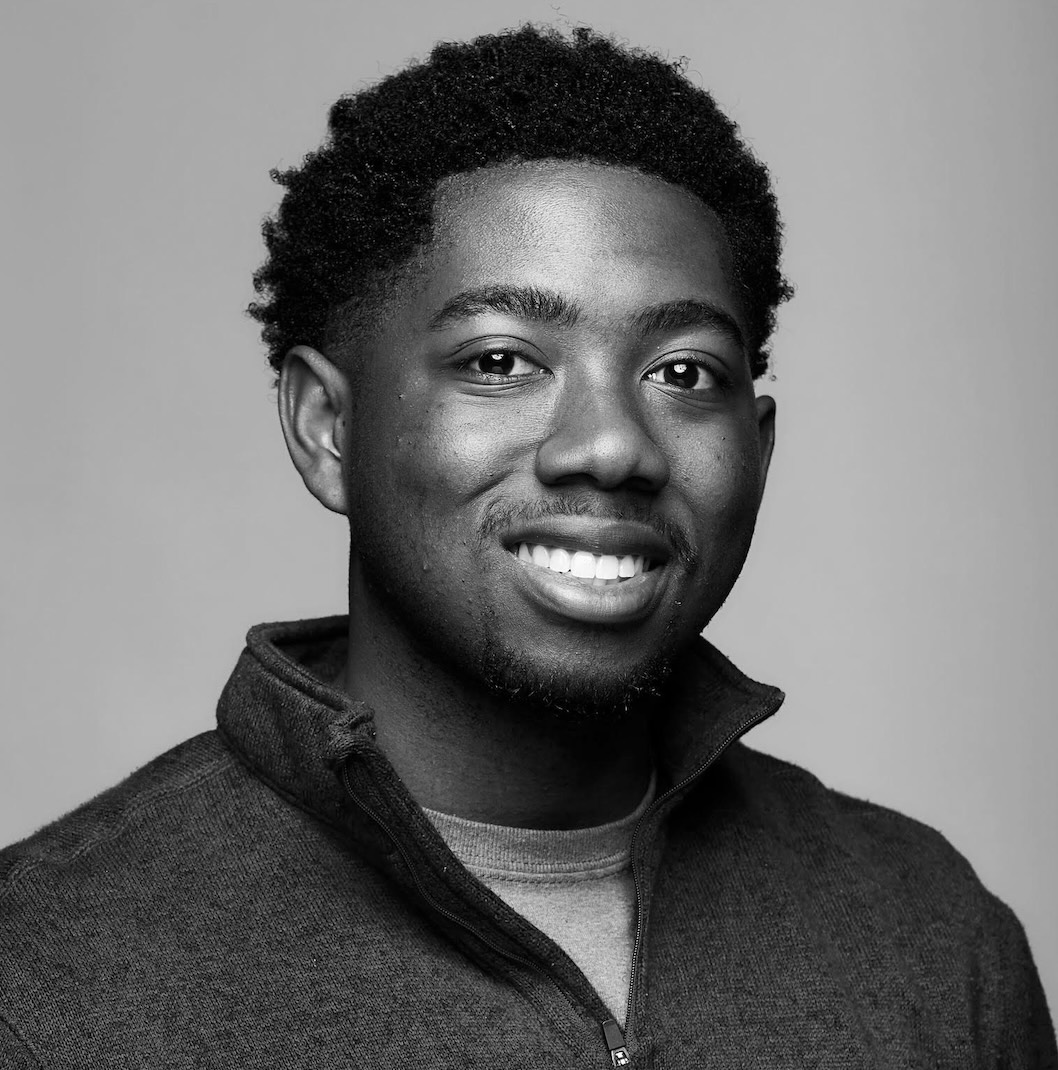}}]{Lenworth Thomas}
received his Bachelor's in Science in Mechanical Engineering from the University of Florida in 2021. He is currently a Ph.D. candidate at Carnegie Mellon University studying Mechanical Engineering with a focus in Robotics.
His research interests include design of robotics systems, robotics for environmental data collection, sensor fusion, autonomous navigation, and field robotics.
\end{IEEEbiography}

\begin{IEEEbiography}[{\includegraphics[width=1in,height=1.25in,clip,keepaspectratio]{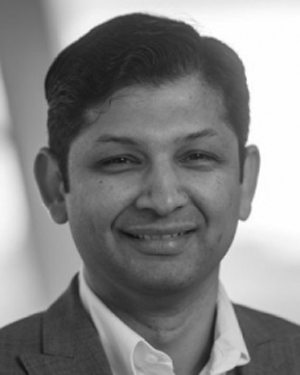}}]{Karthik Dantu}
is an Associate Professor in Computer Science and Engineering at University at Buffalo, State University of New York. Prior to that, he was a Postdoctoral Fellow in EECS at Harvard from 2010-13. He received his M.S. and Ph.D. in Computer Science from University of Southern California in 2010. Karthik is the founding director of the Center for Embodied Autonomy and Robotics at University at Buffalo, a university-wide center for robotics research. His research interests are in robot perception, mobile perception and safe and trustworthy autonomy. 
\end{IEEEbiography}
\begin{IEEEbiography}[{\includegraphics[width=1in,height=1.25in,clip,keepaspectratio]{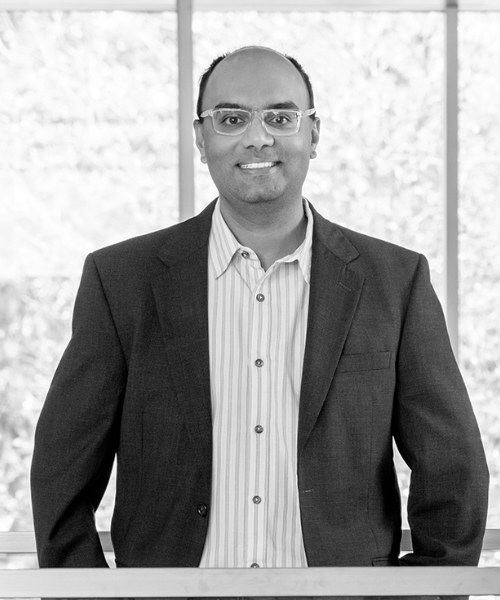}}]{Sanjeev J. Koppal}
is an Associate Professor at the University of Florida’s Electrical and Computer Engineering Department and is a Kent and Linda Fuchs Faculty Fellow. He also held a UF Term Professorship for 2021-23. Sanjeev is the Director of the FOCUS Lab at UF. Since 2022, Sanjeev has been an Amazon Scholar with Amazon Robotics. Prior to joining UF, he was a researcher at the Texas Instruments Imaging R\&D lab. Sanjeev obtained his Masters and Ph.D. degrees from the Robotics Institute at Carnegie Mellon University. After CMU, he was a postdoctoral research associate in the School of Engineering and Applied Sciences at Harvard University.
\end{IEEEbiography}

\end{document}